\documentclass{article}
\usepackage[T1]{fontenc}
\usepackage{preprint,times}
\usepackage{amsmath,amssymb}
\usepackage{booktabs}
\usepackage{array}
\usepackage{longtable}
\usepackage{graphicx}
\usepackage{xcolor}
\usepackage{tikz}
\usetikzlibrary{arrows.meta, positioning, fit, backgrounds, matrix}
\definecolor{okverm}{HTML}{D55E00}\definecolor{okgreen}{HTML}{009E73}\definecolor{okblue}{HTML}{0072B2}
\newcommand{\figval}[1]{\mbox{\textcolor{okblue}{\textbf{#1}}}}
\newcommand{\tightpar}{\looseness=-1\relax}
\newcommand{\resultshead}[2]{\paragraph{#2}\label{#1}}
\tikzset{
  flame/.pic={
    \fill[okverm] (0,0) .. controls (-0.12,0.02) and (-0.11,0.16) .. (0.0,0.29) .. controls (0.03,0.19) and (0.12,0.15) .. (0.09,0.04) .. controls (0.07,0.0) and (0.03,-0.01) .. (0,0) -- cycle;
    \fill[orange!45!yellow] (0.005,0.03) .. controls (-0.05,0.05) and (-0.04,0.11) .. (0.012,0.17) .. controls (0.02,0.11) and (0.055,0.085) .. (0.04,0.04) -- cycle;
  },
  snowflake/.pic={
    \foreach \a in {0,60,120} {\draw[okblue, line width=0.55pt] (\a:0.125) -- (\a+180:0.125);}
    \foreach \a in {0,60,...,300} {\draw[okblue, line width=0.4pt] (\a:0.075) -- ++(\a+40:0.045) (\a:0.075) -- ++(\a-40:0.045);}
  }
}
\usepackage{hyperref}
\hypersetup{colorlinks=true, linkcolor=red!50!black, citecolor=blue!50!black, urlcolor=blue!80!black}
\usepackage{url}
\usepackage{placeins}
\newif\ifledger
\ledgerfalse

\title{Asking for What Was Never Requested:\\ Horizontal and Vertical\\ Proactivity in Agents}

\author{Ido Levy\textsuperscript{1,2}, Asaf Yehudai\textsuperscript{1}, Segev Shlomov\textsuperscript{1}, Asaf Adi\textsuperscript{1}, Leshem Choshen\textsuperscript{1,2}\\
\textsuperscript{1}IBM, \textsuperscript{2}Weizmann Institute of Science\\[3pt]
\href{https://dolev31.github.io/ProactiveInquirer/}{\raisebox{-0.125em}{\includegraphics[height=1em]{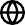}}\hspace{0.3em}Project page}\hspace{1.5em}%
\href{https://github.com/dolev31/ProactiveInquirer}{\raisebox{-0.125em}{\includegraphics[height=1em]{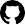}}\hspace{0.3em}dolev31/ProactiveInquirer}\hspace{1.5em}%
\href{https://huggingface.co/dolev31/ProactiveInquirer-Qwen3-8B}{\raisebox{-0.125em}{\includegraphics[height=1em]{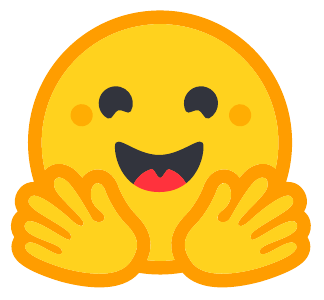}}\hspace{0.3em}dolev31/ProactiveInquirer-Qwen3-8B}}

\prefinalcopy
\extrafloats{100}

\hypersetup{
  pdftitle={Asking for What Was Never Requested: Horizontal and Vertical Proactivity in Agents},
  pdfauthor={Ido Levy, Asaf Yehudai, Segev Shlomov, Asaf Adi, Leshem Choshen},
  pdfsubject={What a tool-using LLM agent should pursue that the user never asked for: horizontal and vertical proactivity, need graphs, and Q\&D (questioner and drafter)},
  pdfkeywords={proactive agents, proactivity, LLM agents, tool-using agents, information seeking, question asking, clarifying questions, need graph, multi-hop question answering, agentic search, preference optimization, DPO, tau2-bench},
  pdflang={en-US}
}
\begin{document}

\maketitle

\begin{abstract}
An agent that uses tools typically responds to what the user explicitly asks, yet completing the task
may require information the user never requested. Work on proactive agents mainly studies whether and
when an agent should act on its own, not what information it should pursue. We study a distinct axis of
proactivity: its content. Horizontal proactivity pursues unstated information that the current context
already identifies, and vertical proactivity pursues needs that only earlier evidence reveals. A need
graph, recovered from a benchmark's own decomposition, records which needs depend on which, so both
forms, and whether the agent stops at the right time, can be scored from a transcript without a model
judge. To learn this behavior, we propose Q\&D (questioner and drafter), which trains a questioner to prefer the question whose
continuation retrieves more of the required evidence, with no reward model or judge. On held-out splits
of three multi-hop question-answering benchmarks, at equal retrieval spend, the trained questioner
improves both forms of proactivity over the same model, prompted, and outperforms a prompted model
$15\times$ larger in the same role on two of the three, and the gain persists after controlling for
question volume and length. Without further training, we place the questioner in an interactive
customer-service agent with a simulated customer, where it completes more tasks while asking fewer
questions, and in retail it outperforms the $15\times$ larger model with fewer follow-up turns from the
customer. These results show that proactivity depends not only on whether an agent acts without being
asked, but also on what it chooses to pursue and when it stops.\tightpar
\end{abstract}

\section{Introduction}
\label{sec:intro}
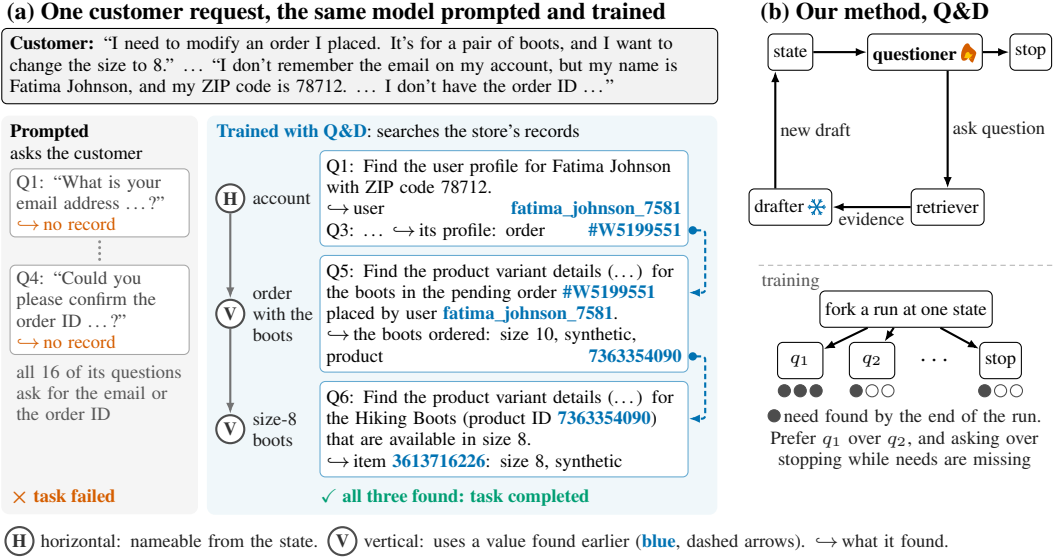
\begin{figure}[t]
\centering
\begin{tikzpicture}[
  font=\footnotesize,
  box/.style={draw, rounded corners=2pt, minimum height=4.8mm, inner sep=2pt, fill=white, align=center, font=\scriptsize},
  arr/.style={-{Latex[length=1.5mm]}, thick},
  hb/.style={circle, draw=black!70, thick, text=black!80, fill=white, font=\scriptsize\bfseries, inner sep=0.3pt, minimum size=3.7mm},
  ttl/.style={font=\footnotesize\bfseries, anchor=west},
  note/.style={font=\scriptsize, align=left, text=black!88, execute at begin node={\hyphenpenalty=10000\exhyphenpenalty=10000}},
  dot/.style={circle, draw=black!70, inner sep=0pt, minimum size=1.6mm},
  full/.style={dot, fill=black!70},
  qstep/.style={font=\scriptsize, align=left, inner sep=2.4pt, rounded corners=2pt, anchor=north west, fill=white, execute at begin node={\hyphenpenalty=10000\exhyphenpenalty=10000}},
  pstep/.style={qstep, draw=black!35, text=black!72, text width=2.2cm},
  tstep/.style={qstep, draw=okblue!60, text width=4.7cm},
  head/.style={font=\scriptsize, anchor=north west, inner sep=0pt, align=left},
  lane/.style={rounded corners=3pt, inner sep=1.1mm}
]
\node[ttl] at (-0.1,3.12) {(a) One customer request, the same model prompted and trained};
\node[draw, rounded corners=2pt, fill=black!5, text width=9.26cm, align=left, inner sep=3pt, font=\scriptsize, anchor=north west, execute at begin node={\hyphenpenalty=10000\exhyphenpenalty=10000}] (req) at (-0.05,2.92)
  {\textbf{Customer:} ``I need to modify an order I placed. It's for a pair of boots, and I want to change the size to 8.'' \ldots\ ``I don't remember the email on my account, but my name is Fatima Johnson, and my ZIP code is 78712. \ldots\ I don't have the order ID \ldots''};
\node[head] (ph) at ([yshift=-2.6mm, xshift=1.1mm] req.south west) {\textbf{Prompted}\\asks the customer};
\node[head] (th) at ([xshift=2.73cm] ph.north west) {\textbf{\textcolor{okblue}{Trained with Q\&D}}: searches the store's records};
\node[pstep] (p1) at ([yshift=-1.2mm] ph.south west) {Q1: ``What is your email address \ldots?''\newline\textcolor{okverm}{$\hookrightarrow$\,no record}};
\node[pstep] (p2) at ([yshift=-3.6mm] p1.south west) {Q4: ``Could you please confirm the order ID \ldots?''\newline\textcolor{okverm}{$\hookrightarrow$\,no record}};
\draw[black!45, line width=1.1pt, dash pattern=on 0pt off 2.1pt, line cap=round, shorten >=1.6pt, shorten <=1.6pt] (p1.south) -- (p2.north);
\node[note, text=black!65, text width=2.2cm, anchor=north west, inner sep=2.4pt] (pn) at ([yshift=-0.4mm] p2.south west) {all 16 of its questions ask for the email or the order ID};
\node[tstep] (t1) at ([xshift=1.37cm, yshift=-1.2mm] th.south west) {\strut Q1: Find the user profile for Fatima Johnson\newline with ZIP code 78712.\newline$\hookrightarrow$\,user\hfill\figval{fatima\_johnson\_7581}\newline Q3: \ldots\ $\hookrightarrow$\,its profile: order\hfill\figval{\#W5199551}};
\node[tstep] (t2) at ([yshift=-1.1mm] t1.south west) {\strut Q5: Find the product variant details (\ldots) for\newline the boots in the pending order \figval{\#W5199551}\newline placed by user \figval{fatima\_johnson\_7581}.\newline$\hookrightarrow$\,the boots ordered: size 10, synthetic,\newline product\hfill\figval{7363354090}};
\node[tstep] (t3) at ([yshift=-1.1mm] t2.south west) {\strut Q6: Find the product variant details (\ldots) for\newline the Hiking Boots (product ID \figval{7363354090})\newline that are available in size 8.\newline$\hookrightarrow$\,item \figval{3613716226}: size 8, synthetic};
\foreach \i/\f/\n in {1/H/account, 2/V/order\\with the\\boots, 3/V/size-8\\boots} {
  \node[hb] (h\i) at ([xshift=0.19cm] th.west |- t\i.west) {\f};
  \node[note, anchor=west, inner sep=0pt] at ([xshift=0.9mm] h\i.east) {\n};
}
\draw[arr, black!55] (h1) -- (h2);
\draw[arr, black!55] (h2) -- (h3);
\node[font=\scriptsize\bfseries, text=okgreen, anchor=north west, inner sep=0pt] (tend) at ([yshift=-1.5mm] t3.south west) {\checkmark\ all three found: task completed};
\node[font=\scriptsize\bfseries, text=okverm, anchor=north west, inner sep=0pt] (pend) at (p2.west |- tend.north) {$\boldsymbol{\times}$ task failed};
\coordinate (gut) at ([xshift=2.7mm] t1.east);
\begin{scope}[on background layer]
\node[lane, fill=black!4, fit=(ph)(p1)(p2)(pn)(pend)] (pl) {};
\node[lane, fill=okblue!6, fit=(th)(h1)(t1)(t2)(t3)(tend)(gut)] (tl) {};
\end{scope}
\begin{scope}[okblue, line width=0.8pt, dash pattern=on 2pt off 1.2pt, rounded corners=1.2pt, {Circle[length=1.1mm]}-{Latex[length=1.5mm]}]
\draw ([yshift=-30.2pt] t1.north east) -- ++(2.2mm,0) |- ([yshift=-14.2pt] t2.north east);
\draw ([yshift=-38.2pt] t2.north east) -- ++(2.2mm,0) |- ([yshift=-14.2pt] t3.north east);
\end{scope}
\node[note, anchor=north west, text width=13.6cm, inner sep=1pt] at ([yshift=-1.0mm] req.west |- tl.south) {\tikz[baseline=(x.base)]\node[hb](x){H}; horizontal: nameable from the state.\enspace \tikz[baseline=(x.base)]\node[hb](x){V}; vertical: uses a value found earlier (\textcolor{okblue}{\textbf{blue}}, dashed arrows).\enspace $\hookrightarrow$\,what it found.};
\begin{scope}[shift={(9.97,0)}]
\node[ttl] at (-0.1,3.12) {(b) Our method, Q\&D};
\node[box] (st) at (0.42,2.6) {state};
\node[box, font=\scriptsize\bfseries] (q) at (2.2,2.6) {questioner\,\tikz[baseline=0.2ex]\pic[scale=0.95]{flame};};
\node[box] (sp) at (3.6,2.6) {stop};
\node[box] (rt) at (2.5,0.55) {retriever};
\node[box] (dr) at (0.42,0.55) {drafter\,\tikz[baseline=-0.6ex]\pic[scale=0.95]{snowflake};};
\draw[arr] (st) -- (q);
\draw[arr] (q) -- (sp);
\draw[arr] (q.south -| rt.north) -- node[right, note, inner sep=1.5pt] {ask question} (rt.north);
\draw[arr] (rt) -- node[below=1pt, note, inner sep=0.5pt] {evidence} (dr);
\draw[arr] ([xshift=-2mm] dr.north) -- node[right, note, inner sep=1.5pt] {new draft} ([xshift=-2mm] st.south);
\draw[black!30, dash pattern=on 2pt off 1.5pt] (0.0,-0.22) -- (3.9,-0.22);
\node[note, text=black!65, anchor=north west, inner sep=1pt] at (0.0,-0.28) {training};
\node[box] (fs) at (1.95,-0.8) {fork a run at one state};
\node[box, minimum width=6mm] (c1) at (0.55,-1.48) {$q_1$};
\node[box, minimum width=6mm] (c2) at (1.5,-1.48) {$q_2$};
\node at (2.35,-1.48) {$\cdots$};
\node[box] (c3) at (3.2,-1.48) {stop};
\foreach \c in {c1,c2,c3} \draw[arr] (fs) -- (\c);
\foreach \x/\n in {0.55/3, 1.5/1, 3.2/1} {
  \foreach \i in {1,...,3} {
    \pgfmathsetmacro{\px}{\x + (\i - 2)*0.21}
    \ifnum\i>\n \node[dot] at (\px,-1.88) {}; \else \node[full] at (\px,-1.88) {}; \fi
  }
}
\node[note, text width=3.7cm, align=center, anchor=north] at (1.93,-2.0) {\tikz\node[full]{};\,need found by the end of the run. Prefer $q_1$ over $q_2$, and asking over stopping while needs are missing};
\end{scope}
\end{tikzpicture}

\caption{Trained with Q\&D (questioner and drafter), our algorithm, an agent finds for itself what a
prompted agent asks the customer to supply. (a) A retail customer-service task. A dedicated prompted agent
asks for the email and order number, which the customer cannot give, and the boots are never
changed. The same model, trained, looks up the account from the name and ZIP code (horizontal), then the
order and the size-8 boots from values it found (vertical), and completes the task. (b) Q\&D: the trained
questioner (flame) asks one question or stops, and a frozen drafter (snowflake) rewrites the draft.
Training forks a run at one state and prefers the candidate whose continuation finds more of the task's
needs (dots).}
\label{fig:overview}
\end{figure}

A tool-using agent usually does what the user asks, yet a task often needs information the user never
mentions \citep{lu2024proactivebench}. A customer who asks a store's agent to change the boots in a pending
order to size 8, same material, gives a name and ZIP code but not the order or an email they no
longer remember (Figure~\ref{fig:overview}a). An agent that waits for the customer to supply these asks
for the email and order number sixteen times, and the boots are never changed. A
proactive agent instead looks up the account from the name and ZIP code, then
the order holding the boots, then size-8 boots in their material, and completes the task. Pursuing a need the current state already names, like the
account, is horizontal proactivity, and pursuing one that only newly found evidence names, like the order
and then the size-8 boots, is vertical proactivity. Without either, an agent stalls or hands the work back to the user as follow-up
questions.

Work on proactive agents asks whether and when an agent should act on its own
\citep{horvitz1999mixed,lu2024proactivebench,tang2026proagentbench}, and rarely what information it should
seek unasked. Agentic search methods decide what to retrieve next and when to stop, but they
either follow the chain the question itself spells out \citep{press2023selfask,trivedi2023ircot} or are
trained only on whether the final answer is correct \citep{jin2025searchr1,chen2025research}, so what each
question pursues is neither measured nor rewarded. Evaluations can also mistake asking more for asking
better, since many imprecise questions may find as much as a few precise ones
\citep{kapoor2025agents,erol2025costofpass}.\tightpar

We make the content of proactivity measurable with what we call a need graph: the evidence a task
requires, with an edge wherever one need can be named only after another is found. Multi-hop
question-answering benchmarks supply these graphs naturally through their own decompositions. A run is
scored from its transcript against the graph, with no model judge, and agents are compared after the same
number of questions, so asking more cannot pass for asking better.

We propose Q\&D (questioner and drafter), an algorithm that embeds proactive information-seeking in an
agent. It gives the agent a questioner, which asks one question at a time or stops, and a frozen drafter,
which folds evidence into a draft. Because the drafter is frozen, every change in what
the agent holds is caused by a question, so each question can be credited with what followed it. To train
the questioner, we fork a run at one step (Figure~\ref{fig:overview}b), continue it after several candidates, and prefer the question whose continuation finds more required evidence, so a question that reaches an unstated need early wins. While evidence is still missing, asking is preferred
over stopping. Every label comes from what followed a question, so Q\&D needs no reward model or
judge. The questioner never sees a graph, so it can run where none exists.\tightpar

Training on consequences makes the agent proactive in both directions. On held-out tasks from three
multi-hop question-answering benchmarks \citep{trivedi2022musique,geva2021aristotle,ho2020twowiki}, where a
question is a search query over the task's evidence, the trained 8B questioner recovers more of the unstated
evidence than the same model, prompted, both what the current state already names and what only newly found
evidence names. On the benchmark whose questions need the most lookups in sequence, it finds 90\% of the
required evidence against 78\%. The gain comes from what it asks, not from asking more or longer questions,
and the 8B questioner also leads a prompted model $15\times$ larger in the same role on two of three benchmarks. Two problems remain: training teaches what to ask more readily than when to stop, and the extra evidence does not yet reach answers.

Q\&D also generalizes beyond question answering. Without further training, in a customer-service agent
serving a simulated customer \citep{barres2025tau2}, it more than doubles retail task success, and, acting
proactively, it completes more retail tasks than the $15\times$ larger model with fewer follow-up turns from
the customer, doing work the customer would otherwise supply.

We make two contributions.
\begin{itemize}
\item A measurable content axis of proactivity. We define horizontal and vertical proactivity on a need
graph and propose evaluation metrics for both and for stopping.
\item Q\&D, an algorithm that embeds both forms of proactivity in an agent by training its questioner on
what each of its questions goes on to retrieve, with no reward model or judge.
\end{itemize}

\section{A need graph makes the content of proactivity measurable}
\label{sec:axes}

We grade the content of proactivity by where each question goes next. At any point of a run, the frontier
holds the needs the agent can already name but has not resolved. Vertical proactivity goes deeper: it
pursues a need that the need just resolved made nameable, as the account's record names the order holding
the boots. Horizontal proactivity goes wider: it pursues any other need on the frontier, one the request
implied, like the account itself at the start, or one that earlier evidence opened, like another order in
the same account. The two are separable: a policy can resolve every surface need
and follow no chain, or follow one chain and miss the rest.

A need graph makes this measurable. For each task it lists the needs, the units of evidence the task
requires, with a prerequisite edge wherever one need can be named only after another is resolved. A need can open several others, so the graph branches wherever a need does, as
an account opens each of its orders. A need's depth is the length of the longest prerequisite path above
it, so needs at depth zero can be named from the request and deeper ones only after the evidence above
them. A need is resolved when the retriever returns its gold paragraph, matched by identifier. Needs joined by prerequisite edges form one independent line of inquiry. We recover the graphs
mechanically from the decompositions the benchmarks ship (Section~\ref{sec:setup}).

We propose the evaluation metrics in Table~\ref{tab:defs}, each read from a finished run against its graph.
Breadth, the number of independent lines a run advances past their first need, reads horizontal
proactivity, and depth-weighted recall and the deepest need resolved read vertical proactivity. Required-evidence coverage, the recall counterpart of the
supporting-paragraph score of \citet{trivedi2022musique}, reads both, including branches inside a line. Two
rates read stopping over the states each policy reaches: whether it stops once everything is found, and whether it keeps asking while it is not.
Where a user is in the loop, proactivity should also spare them: we count the user's follow-up
turns, read together with task success, since giving up also spares the user. The agent never sees a graph: graphs only
score runs and label training data (Appendix~\ref{app:prompts}).

\begin{table}[t]
\caption{What is read from a run against its need graph. $V$ is the set of required needs, $S_t$ the
evidence held after $t$ questions ($S$ at the end), $d_v$ a need's depth, and $C_d$ the coverage at depth $d$. Appendix~\ref{app:metrics} gives each estimator
and its edge cases.}
\label{tab:defs}
\begin{center}\small
\begin{tabular}{@{}l@{\hspace{8pt}}l@{}}
\toprule
quantity & definition \\
\midrule
required-evidence coverage & share of the required needs recovered, $|V\cap S|/|V|$ \\
breadth & independent lines in which the run resolves a need at depth $\geq 1$ \\
depth-weighted recall & $\sum_d w_d C_d / \sum_d w_d$ with $w_d = 0.5, 1, 1.5, 2$ for $d = 0, 1, 2$ and 3 to 5 \\
deepest need resolved & $\max\{d_v : v\in V\cap S\}$, the level the run reached \\
out-of-order rate & share of resolved needs taken before a prerequisite (lower is better) \\
stop-when-done & $\Pr[\text{stop}\mid V\subseteq S_t]$ over the decision points the policy reaches \\
ask-when-not-done & $\Pr[\text{ask}\mid V\not\subseteq S_t]$ over the same decision points \\
\bottomrule
\end{tabular}
\end{center}
\end{table}

\section{Q\&D trains a questioner on what its questions retrieve}
\label{sec:questioner}

Q\&D splits the agent into a questioner, which decides what to ask and when to stop, and a drafter,
which keeps the answer (Figure~\ref{fig:overview}b). At each step the questioner sees the state:
the task, the evidence so far, the current draft and its past questions. It then asks one
question, which a retriever answers from the task's evidence pool, or stops. A frozen
drafter rewrites the draft from the evidence. At the end a frozen answerer, the same in
every arm and held to a fixed word cap, writes the final answer from the task, the evidence and the draft,
so only the questioner differs between policies, and answer length favors none of them. The draft exposes the frontier: once a need's
prerequisite is in the draft, the need can be named, so at every step the questioner chooses between going
deeper along the chain the last evidence opened and opening a need already nameable beside it.

Because the drafter is a fixed function of the evidence, every change in the state is caused by a
question, so Q\&D labels each decision by its consequences, not its wording. A recorded run is forked at a step,
eight alternative questions are sampled there, and each is continued to the end by the policy that produced
the run (Figure~\ref{fig:overview}b, bottom). The candidates share the task, the evidence and the history, so they differ only in the question
asked. Pairs are ordered by consequence alone: first by whether the run answered the task, which decides 6\%
of the pairs we train on, then by which reached the complete evidence
sooner, then by how much evidence each turn added. The preferred question is thus usually the more
proactive one, which reaches the needs the request left unstated sooner, whether by going deeper or wider.
At a state the need graph marks unfinished, asking is ranked above stopping. Training needs tasks whose required
evidence and prerequisites are known, as multi-hop benchmarks provide, but the trained questioner reads no
graph, which is why it can serve a customer-service agent that has none.

The questioner is trained in three stages. It first imitates good decisions: where the required evidence
was already in hand the target is to stop, and where it was not, the best sampled question if its
consequence score clears a fixed floor (Appendix~\ref{app:training}). It is then trained by direct preference
optimization \citep{rafailov2023dpo} on question pairs, and last on
question pairs and stop contrasts together. Stop contrasts rank asking above stopping
at unfinished states, the lesson imitation cannot give, because it writes no target where no sampled
question clears the floor (Section~\ref{sec:stopping}).

\section{Experimental setup}
\label{sec:setup}

Training data are mined from MuSiQue \citep{trivedi2022musique}, StrategyQA \citep{geva2021aristotle}
and 2WikiMultiHopQA \citep{ho2020twowiki}, whose need graphs cover $800$, $2{,}290$ and $12{,}576$ tasks.
Results are read on held-out
test splits of $200$ tasks per benchmark, which we call suites, at two rollout seeds each, and the few read on development tasks say so.
MuSiQue carries the largest share of needs at depth two or more, StrategyQA the decompositions its questions leave implicit,
and 2WikiMultiHopQA the only natural branching. Checkpoints were selected on $132$ MuSiQue and $333$ StrategyQA development tasks
(Appendix~\ref{app:training}).
FRAMES, which ships no need graph, is scored on whether the gold answer or an alias
appears in the answer \citep{krishna2025frames}, and transfer
is read on the retail and airline domains of $\tau^2$-bench \citep{barres2025tau2}, neither used for
training.

Need graphs are recovered mechanically from structure each benchmark ships, so no person and no model
wrote a need or an edge. MuSiQue writes a later step as, for example,
``What is the birthplace of \#1?'', its authors' statement that the step waits on step one. StrategyQA
carries the same references on $2{,}999$ of its $6{,}720$ steps, and 2WikiMultiHopQA links a need to the
one whose subject is its object (Appendix~\ref{app:metrics}). Depth here is therefore depth in the
benchmark's decomposition.\tightpar

The questioner is Qwen3-8B with a low-rank adapter \citep{yang2025qwen3,hu2021lora}, and we report two
training seeds of its final stage. The main comparator is the same model, prompted with the same template.
Beside it we run GPT-OSS-120B, a model $15\times$ larger, prompted \citep{openai2025gptoss}, Claude Opus 5
prompted plainly, a baseline that never asks, and the structured retrieval algorithm PAR$^2$-RAG
\citep{li2026par2rag}, reimplemented with our retriever and drafter and run on Qwen3-8B, GPT-OSS-120B and
Claude Opus 5. The drafter and the answerer are GPT-OSS-120B in every arm except the single-model control of
Section~\ref{sec:controls}. Retrieval is BM25 over each task's released pool of 20 paragraphs (10 on
2WikiMultiHopQA), returning the top 5, 2 and 3 on MuSiQue, StrategyQA and 2WikiMultiHopQA deterministically,
and cost is counted in retrieval calls, one per question.\tightpar

Two readings are kept apart. Equal spend reads both policies, on each task, at the lower of their two
question counts: since the questioner is never told a budget, its first $k$ questions are the ones a
budget of $k$ would have produced, so the reading is one of evidence per question. Own stop, the second
reading, lets each policy decide for itself when to stop, with at most eight calls, and reads it where it
stopped. Each contrast is paired at the task, whose value averages its two rollout seeds and, for the
trained questioner, its two training seeds, and intervals are bias-corrected and accelerated bootstraps
over tasks \citep{efron1987bca}, printed at $10{,}000$ resamples. We call a cell decided when its intervals
from three independent $50{,}000$-resample bootstrap runs all exclude zero, a guard against resampling
error rather than a multiplicity correction, and no question-answering contrast was registered in advance
as a confirmatory test. Each comparator is read at its own lower count, so contrasts do not subtract.

\section{Results}
\label{sec:results}

\resultshead{sec:axes-results}{Q\&D recovers more of the required evidence, and more of it deep.}
\begin{table}[t]
\caption{At equal spend, the trained questioner recovers more of the required evidence than the same model,
prompted, on every suite, and more than GPT-OSS-120B on two of three. Each column pair gives a baseline's
level and Q\&D's over $200$ held-out tasks per suite, the baseline being the same model, prompted, except on
rows that name GPT-OSS-120B. At equal spend a pair is read at its lower question count, so Q\&D's level
varies by comparator. Bold marks the better of a pair where the difference is decided
(Section~\ref{sec:setup}), and $\downarrow$ a measure where lower is better. Differences in the text are
computed before rounding (Table~\ref{tab:app-headline-diffs}).}
\label{tab:heldout}
\label{tab:regimes}
\begin{center}\small
{\renewcommand{\arraystretch}{0.9}%
\begin{tabular}{@{}l@{\hspace{8pt}}w{c}{34pt}@{\hspace{4pt}}w{c}{34pt}@{\hspace{10pt}}w{c}{34pt}@{\hspace{4pt}}w{c}{34pt}@{\hspace{10pt}}w{c}{34pt}@{\hspace{4pt}}w{c}{34pt}@{}}
\toprule
 & \multicolumn{2}{c}{MuSiQue} & \multicolumn{2}{c}{StrategyQA} & \multicolumn{2}{c}{2WikiMultiHopQA} \\
\cmidrule(lr){2-3}\cmidrule(lr){4-5}\cmidrule(l){6-7}
 & Baseline & Q\&D & Baseline & Q\&D & Baseline & Q\&D \\
\midrule
\multicolumn{7}{@{}l}{\textit{Equal spend}} \\
Required-evidence coverage (\%) & 78.3 & \textbf{89.5} & 78.4 & \textbf{85.5} & 88.2 & \textbf{93.1} \\
Depth-weighted recall (\%) & 77.9 & \textbf{90.5} & 51.2 & \textbf{56.6} & 84.8 & \textbf{90.3} \\
Deepest need resolved (depth) & 1.53 & \textbf{1.75} & 0.54 & \textbf{0.64} & 0.57 & 0.63 \\
Breadth (lines) & 0.87 & \textbf{0.99} & 0.47 & \textbf{0.52} & 0.77 & \textbf{0.84} \\
Out-of-order rate (\%) $\downarrow$ & \textbf{17.1} & 22.8 & 11.5 & 12.0 & 4.4 & 3.2 \\
Coverage vs GPT-OSS-120B (\%) & 77.8 & \textbf{84.8} & 80.7 & \textbf{85.0} & \textbf{92.4} & 88.9 \\
\midrule
\multicolumn{7}{@{}l}{\textit{Own stop (at most eight calls)}} \\
Required-evidence coverage (\%) & 81.6 & \textbf{91.2} & 87.6 & 86.4 & 94.6 & 96.1 \\
Answer token F1 (\%) & 41.1 & 43.6 & 43.5 & 43.1 & 61.4 & 64.5 \\
Questions per task & 5.9 & 4.2 & 6.0 & 3.0 & 3.0 & 2.3 \\
Coverage vs GPT-OSS-120B (\%) & 81.1 & \textbf{91.2} & \textbf{90.0} & 86.4 & 96.8 & 96.1 \\
Questions vs GPT-OSS-120B & 5.3 & 4.2 & 6.1 & 3.0 & 3.1 & 2.3 \\
\bottomrule
\end{tabular}}%

\end{center}
\end{table}

Q\&D makes the agent more proactive on every held-out suite. After the same number of questions, the
trained questioner recovers more of each task's required evidence than the same model, prompted, by 11.2
percentage points on MuSiQue, 7.0 on StrategyQA and 5.0 on 2WikiMultiHopQA (Table~\ref{tab:heldout}). The
gain is stable: each training seed alone is decided on all three suites, and all four
training runs agree within a standard deviation of 0.8 points (Appendix~\ref{app:seedzero}). The
first training stage alone (imitation, Section~\ref{sec:questioner}) also raises coverage at every base size
from 1.7B to 8B, on development tasks (Appendix~\ref{app:scale}).\tightpar

Much of the extra evidence is deep, which is vertical proactivity: depth-weighted recall rises by 12.5, 5.4
and 5.5 points, most on MuSiQue, whose chains run deepest. The deepest need a run resolves also lies deeper,
by 0.22 levels on MuSiQue and 0.10 on StrategyQA (Table~\ref{tab:heldout}). Breadth rises as well, so the
questioner goes deeper without narrowing its search. We attribute the depth gain to the labels, which prefer
the question whose continuation reaches all the required evidence sooner and so favor starting a chain early
enough for later questions to follow it.

Where a task needs both directions at once, the questioner is more proactive in both. On synthetic tasks
built from two to four independent lines of inquiry, chains of needs independent of one another,
each several steps deep, it advances more of the lines than the same model, prompted, when a task has two
or three lines, ties with four, and recovers more of the deep evidence in every case
(Appendix~\ref{app:metrics}). On real data, when two held-out MuSiQue questions are joined into one task
whose answer needs both chains, it advances more of the two chains past their first step and recovers 15.2
points more of the required evidence, as its gains on each question alone predict
(Appendix~\ref{app:composed}).

One side effect appears on MuSiQue: the trained questioner more often finds a deep piece of evidence before
the one it depends on, by 5.7 points (the out-of-order rate in Table~\ref{tab:heldout}). It comes from
paragraphs retrieved by chance, not from what the questions ask for: the questions themselves skip ahead no
more often than the prompted model's, by 1.2 points spanning zero (Appendix~\ref{app:heldout}).

Better questions also make the evidence cheaper. For each required need it finds, the trained questioner
spends 40\% fewer tokens on MuSiQue, 11,610 against 19,354, and fewer on the other two suites as well
(Appendix~\ref{app:matchedcost}). This ratio needs no matching of spend, so the gain does not come from
how spend is matched. We attribute the saving to proactive questioning: each question the
trained questioner asks recovers more of the required evidence.

\resultshead{sec:controls}{The gain comes from what Q\&D asks.}\label{sec:length}

What the questions say matters, not their number. Replacing the trained questioner's questions with
as many drawn at random from other tasks of the same benchmark loses 34.7 points of coverage on MuSiQue
and 63.3 on StrategyQA (Appendix~\ref{app:controls}).\tightpar

The trained questioner's later questions build on what earlier ones found, which is vertical proactivity at
work: if its questions ignored what it read, hiding the evidence would cost nothing. On MuSiQue, removing
the evidence from its state costs 8.4 and 8.7 points of coverage, and letting it plan its questions from the
request alone costs 4.3 and 6.4 (per training seed). On StrategyQA neither makes a detectable
difference, because there the ablated questioners ask little more than one question.

The gain comes from training the questioner, not from splitting the agent in two. When the trained
questioner also writes the draft itself, it stays within a 5-point equivalence margin: equivalent at its own
stop, and 2.2 points behind the split at equal spend
(Appendix~\ref{app:controls}).

Longer questions do not explain the gain either. The trained questioner's questions are longer, 17.5 words
against 13.1 on MuSiQue, but when the prompted model may spend as many question tokens, it still recovers
14.9 and 4.7 points less on MuSiQue and StrategyQA development tasks (Appendix~\ref{app:matchedcost}). The
advantage lies in what the trained questioner asks, not in how much it writes.

People recognize these proactive questions but differ on which is best. Two human raters agree on 88\% of
candidate questions on whether they reach a need the request never states (Cohen's
$\kappa = 0.76$), but on only 68\% of the pairs both judged about which question is the better move
($\kappa = 0.24$), and each agrees with our rule on 65\% and 68\% of pairs, against 50\% by chance
(Appendix~\ref{app:validity}). Here, a better question is one that recovers more of the required evidence,
not one people prefer.

\begin{figure}[t]
\centering
\includegraphics[width=0.88\linewidth]{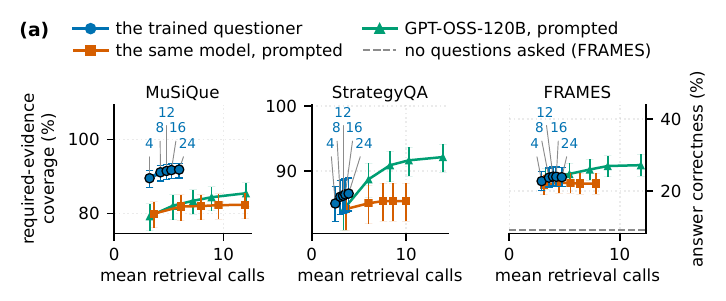}\\[2pt]
\includegraphics[width=0.88\linewidth]{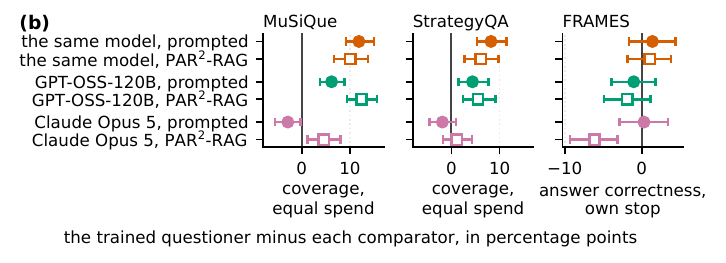}
\caption{On MuSiQue the trained questioner gets more evidence from fewer calls, and a hand-designed
retrieval algorithm does not close the gap. (a) Coverage on MuSiQue and StrategyQA and answer correctness
on FRAMES against the calls used on average, when each policy stops on its own under a budget of 4, 8, 12,
16 or 24 calls (numbered on the trained questioner's points). (b) The trained questioner minus three
models, prompted plainly (filled) or running the structured retrieval algorithm PAR$^2$-RAG
\citep{li2026par2rag} (hollow), in percentage points: coverage at equal spend and FRAMES answers at each
policy's own stop. Bars are bootstrap intervals over tasks.}
\label{fig:frontier}
\label{fig:comparators}
\end{figure}

\resultshead{sec:frontier}{Q\&D's 8B questioner leads one $15\times$ larger on two of three suites.}
Training a small questioner can beat scale (Figure~\ref{fig:frontier}). At equal spend, the trained 8B questioner recovers more of the
required evidence than GPT-OSS-120B, prompted, a model $15\times$ larger in the same role, on MuSiQue and
StrategyQA, by 7.0 and 4.3 points, and trails it on 2WikiMultiHopQA, by 3.5 points (Table~\ref{tab:regimes}).
The lead follows depth: it is largest on MuSiQue, whose chains run deepest, and reversed on
2WikiMultiHopQA, whose needs lie at most one step deep, which suggests that learned proactivity pays most
where evidence must be followed step by step.

The trained questioner's advantage is efficiency, which call budgets make visible (Figure~\ref{fig:frontier}a).
On MuSiQue it leads both prompted models at every budget from 4 to 24 calls, and allowed only 4 calls it
recovers 4.1 points more of the required evidence than GPT-OSS-120B allowed 24, using about a quarter of
the calls (Appendix~\ref{app:stopping}). Elsewhere the prompted models catch up at larger budgets: on StrategyQA the trained questioner ties the same model with fewer questions and trails
GPT-OSS-120B once it spends more, and on FRAMES it trails GPT-OSS-120B at budgets of 16 and 24.

A hand-designed alternative does not close the gap either. PAR$^2$-RAG \citep{li2026par2rag}, a structured
retrieval algorithm that plans sub-questions and checks whether the evidence suffices, never beats plain
prompting of its own model at equal spend on any of three models (Figure~\ref{fig:comparators}b,
Appendix~\ref{app:matchedcost}), and the trained questioner leads it on every model on MuSiQue. The one
comparator ahead of the trained questioner at equal spend is Claude Opus 5, prompted plainly, on MuSiQue, by
2.9 points. On FRAMES, when each policy stops on its own, the one decided difference is that PAR$^2$-RAG
run with Claude Opus 5 answers better, by 6.1 points, while asking more questions, 4.5 against 3.6.

\resultshead{sec:ownstop}{Stopping on its own, Q\&D asks fewer questions and still leads on the deepest chains.}
When each agent decides for itself when to stop, with at most eight calls (own stop), the trained
questioner asks fewer questions than either prompted model on every suite and keeps its lead on MuSiQue,
by 9.6 points over the same model, prompted, and 10.1 over GPT-OSS-120B (Table~\ref{tab:regimes}). On the
other two suites it does not differ detectably from the same model, and on StrategyQA it trails
GPT-OSS-120B by 3.6 points, while that model asks twice as many questions.

The trained questioner stops early by choice, not because it runs out of calls. It uses its whole budget in
only 34\% of MuSiQue runs when allowed 4 calls and in 8\% when allowed 24, against 13\% to 69\% for the
prompted questioners across both suites and budgets (Appendix~\ref{app:stopping}).

\resultshead{sec:stopping}{What each stage of Q\&D's training teaches, and why stopping lags.}
\begin{table}[t]
\caption{What each training signal teaches, on development tasks ($132$ MuSiQue and $333$ StrategyQA), each
row training the same model on one signal (Section~\ref{sec:questioner}), the final stage on both. Gain is
coverage over the same model, prompted, in points at a matched spend (Appendix~\ref{app:matchedcost}).
Stop\,$|$\,done and ask\,$|$\,not done are the percent of decision points at which a policy stops once the
required evidence is in and asks while it is not, and asks counts questions per task. Bold marks each
column's largest value, not a tested difference, and no row leads on all three (intervals in
Appendix~\ref{app:stopping}).}
\label{tab:mechanism}
\begin{center}\small
\begin{tabular}{@{}l@{\hspace{8pt}}c@{\hspace{5pt}}c@{\hspace{5pt}}c@{\hspace{5pt}}c@{\hspace{10pt}}c@{\hspace{5pt}}c@{\hspace{5pt}}c@{\hspace{5pt}}c@{}}
\toprule
 & \multicolumn{4}{c}{MuSiQue} & \multicolumn{4}{c}{StrategyQA} \\
\cmidrule(lr){2-5}\cmidrule(l){6-9}
supervision & gain & stop\,$|$\,done & ask\,$|$\,not done & asks & gain & stop\,$|$\,done & ask\,$|$\,not done & asks \\
\midrule
same model, prompted & --- & 15.3 & 96.4 & 6.29 & --- & 14.6 & 96.3 & 5.59 \\
\midrule
imitation (first stage) & $+4.9$ & 95.1 & 86.8 & 3.05 & $+8.0$ & 90.7 & 83.0 & 1.34 \\
question pairs & $\boldsymbol{+16.7}$ & \textbf{98.7} & 86.0 & 2.47 & $\boldsymbol{+8.4}$ & \textbf{98.0} & 83.4 & 1.28 \\
stop contrasts alone & $-0.5$ & 10.1 & \textbf{99.6} & 6.95 & $-1.7$ & 9.7 & \textbf{98.1} & 5.75 \\
\midrule
final stage, seed 1 & $+15.7$ & 42.7 & 94.2 & 4.23 & $+8.2$ & 54.6 & 91.2 & 2.13 \\
final stage, seed 2 & $+15.4$ & 42.7 & 94.5 & 4.24 & $+7.8$ & 54.5 & 90.6 & 2.14 \\
\bottomrule
\end{tabular}

\end{center}
\end{table}

Each stage of Q\&D's training teaches something different. Imitation, the first stage, copies good decisions, but at an
unfinished state where no sampled question clears the floor it has nothing to copy, so $24{,}538$ such
states drop out of its training data (Appendix~\ref{app:training}), and with them the lesson to keep asking.
The second stage learns from question pairs, which prefer the better of two questions, and the final stage
adds stop contrasts, pairs that prefer asking over stopping at unfinished states. In
Table~\ref{tab:mechanism} the same model is trained on each signal alone: question pairs raise coverage but
stop too early, stop contrasts alone almost never stop, and the final stage, trained on
both, raises coverage about as much while asking when work remains, though it stops half as often as
imitation once the work is done.

On held-out tasks the trained questioner stops more often once the work is done, by 29 to 33 points over
the same model, prompted, on every suite and seed, and by 25, 43 and 12 points over GPT-OSS-120B
(Appendix~\ref{app:stopping}). Where evidence is still missing, it keeps asking as often as the prompted
model on MuSiQue and 2WikiMultiHopQA, but 7 points less often on StrategyQA.

The extra evidence does not yet show in the answers, for two reasons. On MuSiQue the trained questioner finds
the evidence the answer rests on 9.7 points more often, which at the prompted model's answer rates implies
an answer gain of about 4.3 points, below the 4.7 our sample can detect and consistent with the 2.4
observed (Appendix~\ref{app:answerloss}). On StrategyQA it stops before finding that evidence, by choice in
92\% and 86\% (per seed) of the runs that end without it, against 39\% for the prompted model, and
supervising that stop directly does not help (Appendix~\ref{app:stopping}).

\resultshead{sec:transfer}{Q\&D's proactivity generalizes to a customer-service agent.}
\begin{figure}[t]
\centering
\includegraphics[width=\linewidth]{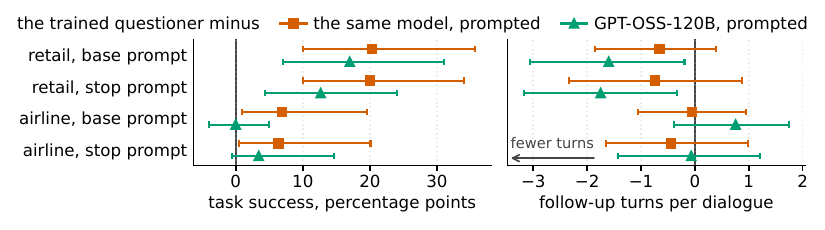}
\caption{In retail, Q\&D's trained questioner completes more $\tau^2$-bench tasks than either prompted
baseline, and needs fewer follow-up turns from the customer than GPT-OSS-120B. Each point is the trained
questioner minus a baseline over $25$ retail or $17$ airline tasks under one of two prompt variants
(Appendix~\ref{app:transfer}): task success in percentage points (left) and the customer's follow-up turns
per dialogue (right). Bars are bootstrap intervals over tasks, each averaging both training seeds.}
\label{fig:tau2}
\end{figure}

Proactivity learned with Q\&D is a general behavior, not a skill tied to question answering. Placed with no
further training in a customer-service agent on $\tau^2$-bench, where a simulated customer brings a task and
each dialogue allows sixteen questions, the trained questioner completes more tasks than the same model,
prompted, in both domains (Figure~\ref{fig:tau2}). In retail, success rises from 13\% and 12\% to 34\% and
32\% under two prompt variants (Appendix~\ref{app:transfer}), better on ten of twenty-five tasks and worse
on none in each, and in airline it rises by 6.9 and 6.4 points. It asks 1.3 and 1.4 fewer questions per
retail dialogue, and more of them reach the records the task needs: it asks less and finds more.

Against a model $15\times$ larger, the agent completes more tasks while the customer supplies fewer turns,
the payoff proactivity promises. GPT-OSS-120B, prompted in the
same role and declared as a baseline before its dialogues ran, completes 17\% and 19\% of retail tasks. The
trained questioner beats it in retail by 17.0 and 12.7 points of success, with 1.6 and 1.8 fewer follow-up turns from the customer, both decided. Against the same model,
prompted, follow-up turns do not differ detectably, and in airline no difference from
GPT-OSS-120B is decided (Appendix~\ref{app:transfer}). We attribute the fewer turns to proactive questioning:
the agent finds more of what the task needs in the store's records.

\section{Related work}
\label{sec:related}

\textbf{Learning what to ask.} The nearest ancestors supervise a question by what it causes, by simulating
the turns that follow it \citep{zhang2025futureturns}, contrasting clarifying with answering
\citep{chen2025act} or pairing terminate-against-continue trajectories \citep{liu2025cart}. Others learn from
counterfactual information gain \citep{kong2026infopo}, causal preferences \citep{zhang2026das}, task success
\citep{yao2026bao}, a judge's verdict or the user effort spared
\citep{acikgoz2025speakrl,sun2026ppp,zhao2026askbench,qian2025userrl}, or self-play with a simulated user
\citep{andukuri2024stargate}. Expected value of information \citep{rao2018learning,aliannejadi2019asking}
needs a model of what an answer will do, where ours needs only the answer, and ranking sampled
continuations by outcome also underlies step-level preference training
\citep{wang2024mathshepherd,lai2024stepdpo}.

\textbf{Proactivity.} Accounts of initiative ask whether and how far an agent acts on its own
\citep{horvitz1999mixed,handler2023taxonomy,SAPKOTA2026103599} and what triggers it
\citep{bui2026proactivity}. The nearest content-side work grades implied needs under
a model judge, without asking whether a need could have been named from the task
\citep{harfi2026proactbench}. Benchmarks that isolate a missing variable, an ambiguous query or a need not
yet stated score whether a model recognizes the gap
\citep{li2025questbench,deng2026interactcomp,tao2026discobench,wu2026atrbench}, and simulated and recorded
users score the timing and form of an intervention
\citep{tang2026proactiveservice,nathani2026pare,tang2026proagentbench,qian2025userbench}. These measure the
decision or its timing. We measure what was pursued.

\textbf{Agentic search and retrieval.} Interleaving reasoning with actions \citep{yao2023react} and
decomposing a question into retrieval steps \citep{press2023selfask,trivedi2023ircot} follow prerequisite
chains by design. Decomposition formalisms write a question
as a graph of dependent steps \citep{wolfson2020break} that prompting can solve in order
\citep{zhou2023leasttomost}. Our need graphs have that shape, but their nodes are the evidence a task requires, and they
only score runs and are never shown to the policy. Policies trained to interleave search with reasoning use outcome rewards
\citep{jin2025searchr1,chen2025research}, and adaptive retrieval learns when to retrieve
\citep{asai2024selfrag,jiang2023flare,jeong2024adaptiverag}. Structured retrieval instead changes the
pipeline, planning and checking sufficiency \citep{li2026par2rag}, synthesizing a program of retrieval calls
\citep{sun2026pyrag}, or curating the pool or a corpus graph
\citep{nahid2025prism,wang2024rear,cahoon2025trex,zhu2026conrag,liu2026a2rag}. Holding the retriever, pool
and drafter fixed, we compare against PAR$^2$-RAG, whose mechanism is a sequence of ask-or-stop decisions.

\textbf{Length and model judges.} Preference training tracks response length
\citep{singhal2023long,liu2024lddpo,dubois2024lengthcontrolled} and model judges carry position and
self-preference bias \citep{zheng2023judging,wataoka2024selfpreference,norman2026reliability}, so the content
claim rests on a length control, no quantity is model-judged, and agreement is
chance-corrected \citep{cohen1960kappa,krippendorff2004content}.\tightpar

\section{Limitations}
\label{sec:limits}

The two reported seeds share one supervised checkpoint and one pair export with two further runs, so
intervals are over tasks alone and do not estimate variation from that checkpoint and export. No checkpoint
met the pre-committed selection rule as written, and the reported configuration was kept on a second ground
fixed in advance (Appendix~\ref{app:training}).
Training labels and the primary quantity share the need graphs, answers decide only 6\% of training pairs,
and answers do not yet move detectably. The released retrieval pools are small enough that most
prerequisite edges do not gate retrieval, so the vertical result concerns depth in a decomposition
(Appendix~\ref{app:validity}), and on MuSiQue and StrategyQA, with one line of inquiry per task, breadth
records only whether a run got past its first need. Candidate questions and rollouts come from language
models, two human raters checked the label rule on one suite, and every user-facing number comes from a
simulated customer. Q\&D is one round of off-policy training, not compared with on-policy reinforcement
learning or iterated on its own rollouts, and the length control was not run held-out. Prompt search and
scale are in Appendices~\ref{app:gepa} and~\ref{app:scale}.

\section{Conclusion}
\label{sec:conclusion}

We introduced the content of proactivity, what an agent pursues unasked, as an axis beside whether and when
it acts, and made its horizontal and vertical forms measurable with a need graph. We
proposed Q\&D, which trains an agent to question proactively, learning from the consequences of its questions, with no reward model or judge. At equal spend, the trained 8B questioner recovers more of the
required evidence than the same model, prompted, reaches deeper into each task's dependencies, and leads a
prompted model $15\times$ larger in the same role on two of three benchmarks. The gain lies in what it asks,
not how much, and is largest where evidence must be followed step by step. It generalizes to a
customer-service agent, completing more tasks with fewer questions, and in retail it outperforms the larger model with fewer follow-up turns from the customer: a proactive
agent does work the user would otherwise supply. Training teaches what to ask more readily than when to
stop, so next are stop supervision that carries the evidence gain into answers and retrieval pools where a
prerequisite must be found before its dependents.

\FloatBarrier

\bibliography{references}
\bibliographystyle{preprint}

\appendix
\section{Metric definitions and estimator}\label{app:metrics}

This section defines the need graph and every quantity of Section~\ref{sec:axes}, reports the
two-axis readings behind Table~\ref{tab:heldout}, and states the populations, eligibility rule and
estimator of Section~\ref{sec:setup}.

\textbf{The need graph.} A task's need graph $N(x)=(V,E)$ holds the units of evidence the task
requires, its needs $V$, with a prerequisite edge $(u,v)\in E$ wherever $v$ becomes nameable only once
$u$ is resolved. The prerequisites of $v$ are $\mathrm{pa}(v)$, its depth $d_v$ is the longest
prerequisite path above it, and the task's facets $\mathcal{F}$ are the weakly connected groups of
prerequisite edges left once the surface needs are removed, a facet being one independent line of
inquiry, the term Table~\ref{tab:defs} uses.

\textbf{How each suite's graph is recovered.} No person and no model wrote any node or edge
(Table~\ref{tab:app-goldcensus}). One pattern recovers MuSiQue's edges from its authors' references
to an earlier step, such as \#1: every one of its $1{,}464$ needs below the surface carries such a
reference and none of the $1{,}196$ on the surface does. StrategyQA carries the
same references on $2{,}999$ of $6{,}720$ steps and leaves the rest unconnected rather than joining
them by position, which would manufacture the structure being measured, so its depth is a lower
bound. 2WikiMultiHopQA links a need to another whose subject is the first's object. In a random-deletion
test on stored runs (Appendix~\ref{app:validity}), a random edge error could hide the depth result but not
manufacture it, and a systematic error is not covered.

\begin{table}[h]
\centering
\caption{The three need-graph sets, every depth recomputed from the stored edges.}
\label{tab:app-goldcensus}
\begin{footnotesize}\setlength{\tabcolsep}{4pt}\begin{tabular}{lrrrll}
\toprule
suite & graphs & nodes & edges & depth histogram & behind an edge \\
\midrule
MuSiQue         &    800 &  2{,}660 & 1{,}860 & 0:1{,}196 \; 1:800 \; 2:530 \; 3:134 & $55.0\%$ \\
StrategyQA      & 2{,}290 &  6{,}720 & 4{,}483 & 0:3{,}721 \; 1:2{,}320 \; 2:615 \; 3:60 \; 4:4 & $44.6\%$ \\
2WikiMultiHopQA & 12{,}576 & 31{,}120 & 11{,}956 & 0:19{,}164 \; 1:11{,}956 & $38.4\%$ \\
\bottomrule
\end{tabular}\end{footnotesize}
\end{table}

\textbf{The quantities and the taxonomy.} A run holds evidence $S_t$ after its $t$th question and $S$
at the end. It resolves $v$ at $t_v=\min\{t: v\in S_t\}$, and $v$ becomes reachable at
$r_v=\min\{t:\mathrm{pa}(v)\subseteq S_t\}$, both $\infty$ if never. Table~\ref{tab:quantities}
defines what is read off this and Table~\ref{tab:taxonomy} arranges the content axis as a ladder.
Coverage at depth two or more weights each depth by the needs it holds, so a shallow frontier of two
needs cannot outvote a deep one of forty. Absent is not zero: a task with nothing at depth two is left
out of the depth-two criterion, and a task with no graph or a suite with no user simulator is omitted
rather than filled in, because a fabricated zero cannot be told from a measurement.

\begin{table}[t]
\caption{Everything this paper measures, read from a finished run against its need graph. Breadth is
a count, since tasks differ in how many lines they have, and a line's facet holds only its needs below
the surface, so breadth counts the lines in which a run resolves a need beyond the line's first.
Depth-weighted recall averages coverage at each depth the task has, $C_d$, with weights $w_d$ of
$0.5$, $1.0$ and $1.5$ at depths $0$, $1$ and $2$ and $2.0$ at depths $3$ to $5$, set in the scorer. A
depth beyond five carries no weight, which matters only on the two deeper constructed corpora, whose
chains run to six and eight. The stopping pair is read at every decision point, excluding states the
budget forced. $\tau$ is the tokens spent.}
\label{tab:quantities}
\begin{center}\footnotesize\setlength{\tabcolsep}{2pt}
\begin{tabular}{@{}l@{\hspace{3pt}}l@{\hspace{3pt}}l@{}}
\toprule
quantity & definition & reads \\
\midrule
required-evidence coverage & $|V\cap S|\,/\,|V|$ & evidence recovered \\
breadth & $|\{f\in\mathcal{F}: f\cap S\neq\emptyset\}|$ & independent lines touched \\
depth-weighted recall & $\sum_{d} w_d\, C_d\,/\,\sum_{d} w_d$, over depths present & coverage per level, weighted deep \\
coverage at depth $k$ & $|\{v: d_v=k,\, v\in S\}|\,/\,|\{v: d_v=k\}|$ & one level on its own \\
deepest need resolved & $\max\{d_v: v\in V\cap S\}$ & the level reached at all \\
out-of-order rate & $|\{v\in S: \exists u\in\mathrm{pa}(v),\; t_u>t_v\}|\,/\,|V\cap S|$ & taken out of order \\
promptness & $|\{v\in S: t_v=r_v\}|\,/\,|\{v\in S: r_v<\infty\}|$ & taken when first reachable \\
stop-when-done & $\Pr[\textsc{stop}\mid |V\cap S_t|=|V|]$ & halts once the evidence is in \\
ask-when-not-done & $\Pr[\textsc{ask}\mid |V\cap S_t|<|V|]$ & continues while evidence is missing \\
cost per need found & $\tau\,/\,|V\cap S|$ & the price of what it found \\
\bottomrule
\end{tabular}
\end{center}
\end{table}

\begin{table}[t]
\caption{The content axis of proactivity as a ladder, each level defined on the need graph. Order and
stopping complete the axis: a policy that reaches deep needs by luck takes them out of prerequisite
order, and knowing that the required evidence is in hand is the content-axis counterpart of staying
silent. Levels 1 to 3 sit one above the depth of the needs they pursue, since level 0 is the request.}
\label{tab:taxonomy}
\begin{center}\footnotesize\setlength{\tabcolsep}{4pt}
\begin{tabular}{@{}cl l l@{}}
\toprule
level & name & the needs it pursues & read by \\
\midrule
0 & stated & named in the request & coverage of stated needs \\
1 & horizontal & at depth zero, unstated & breadth, not separable \\
2 & vertical, one step & after one prerequisite & depth-weighted recall \\
3 & vertical, deep & behind two or more prerequisites & deepest need resolved \\
4 & open & in no annotated graph & not scored by a graph \\
\midrule
   & order & taken before its prerequisite & out-of-order rate (lower better) \\
 & stopping & none, the evidence is in hand & stop-when-done, ask-when-not-done \\
\bottomrule
\end{tabular}
\end{center}
\end{table}

\textbf{Order.} Lower is better on the out-of-order rate: where a step gates an action rather than a
citation, acting out of order is acting on an unconfirmed premise, although no downstream failure is
measured here. Neither it nor promptness separates real vertical
proactivity from luck on the evidence available. Promptness is defined only where each arm's own
trajectory makes it so, which selects its paired population on both arms' behavior, and the
out-of-order rate points the wrong way: on MuSiQue it excludes zero in the adverse direction on the
same runs whose depth-weighted recall rises, a separate measurement and not an offset
(Table~\ref{tab:heldout}, and Appendix~\ref{app:heldout} for where the excess comes from).

\textbf{Why no benchmark suite separates the horizontal axis.} MuSiQue gives every one of its $800$
tasks exactly one facet and StrategyQA none of its $2{,}290$ tasks two or more, so on the two suites
this work trains and selects on, breadth and every variant of it have one bit of range or none. Only
2WikiMultiHopQA branches, on $21.2\%$ of its $12{,}576$ tasks, and there no annotated need sits at
the second level. Nor could a separating suite be assembled from these annotations: no prerequisite
edge crosses a facet boundary in $19{,}532$ edges, and among the $2{,}735$ tasks with two or more
facets none has a facet made entirely of dispensable needs. So on these suites breadth is a
measurement and never an independent axis, and the vertical claim has two controls that could falsify
it (Appendix~\ref{app:controls}) where breadth has none.

\begin{figure}[t]
\centering
\includegraphics[width=\linewidth]{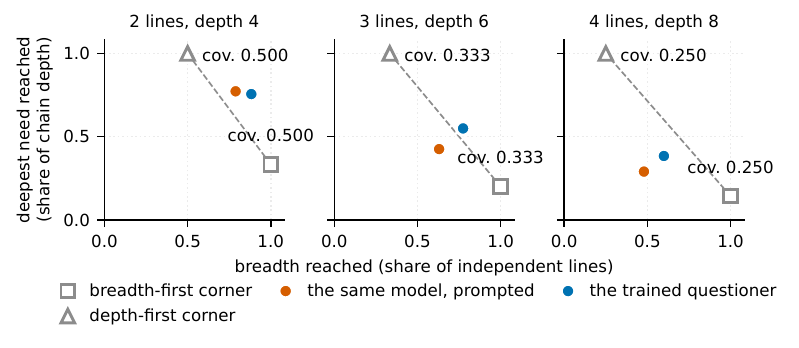}
\caption{The two axes are separable, and the trained questioner reaches further along both than the
same model, prompted, on the two longer chains. Each panel is one constructed corpus, both axes
normalized to what it makes available. A breadth-first corner (square) and a depth-first corner
(triangle) reach identical coverage, labeled at each. Positions are per-arm levels over $400$ runs
each and are not matched on spend: the corners ask exactly
$4$, $6$ and $8$ questions, the prompted arm $9.6$, $8.8$ and $8.5$ and the trained questioner $5.3$,
$7.4$ and $8.4$, so a position relative to the line between the corners reflects spend as well as
policy. The trained questioner asks more than the corners at every shape, yet it sits above the line
at the two shorter chains and below it at the longest, on $8.4$ questions against $8$, so the line is
not an equal-spend frontier and neither side of it ranks an arm against the corners.}
\label{fig:axesseparable}
\end{figure}

\textbf{Constructed corpora, where both axes are live.} Facets exclude the surface, so breadth already
requires depth and the two could have been nested rather than orthogonal. In each of three constructed
shapes the chains run twice as deep as the number of lines, so that two hand-written corner policies
pay equal cost: a breadth-first one takes one hop into every line and a depth-first one follows one
chain. They reach identical coverage within each shape and opposite extremes of breadth and depth
(Figure~\ref{fig:axesseparable}), so coverage cannot separate the axes and breadth and depth can. Being
hand-written, they say nothing about whether a trained policy trades one axis for the other.

\noindent The trained questioner sits in the interior of both axes at every shape, each shape its own
corpus and never pooled. At equal spend against the same model, prompted, it gains breadth at the two
shorter chains, by $+0.215$ $[+0.138, +0.296]$ and $+0.151$ $[+0.046, +0.264]$, ties at the longest,
gains depth-weighted recall on all three, and resolves deeper needs at the middle shape by $+0.354$
$[+0.222, +0.500]$. Each training seed alone decides the same cells. The out-of-order rate takes one
value on all four arms, both corners included, so these corpora cannot express it.

\textbf{On natural questions composed in pairs, both axes are live and the gain carries
over.}\phantomsection\label{app:composed} Two held-out MuSiQue questions that share no template,
title or answer, neither in the training set and at least one with a chain of depth two or more, are
joined into one task that states both, their evidence pools and need graphs unioned with no edge
between them, so each task has exactly two independent lines of inquiry. Both orders of each
pair are separate tasks, giving $48$ pairs and $96$ tasks, with a ceiling of $16$ questions. The
metrics, the prediction and the rule for reporting a negative result were declared before the tasks
were built. At equal realized spend against the same model, prompted, with the bootstrap clustered on
the pair, the trained questioner advances more of the two lines, by $+0.276$
$[+0.172, +0.396]$, covers more of the required evidence, by $+0.152$ $[+0.112, +0.197]$, and weights
more of it toward deep needs, by $+0.159$ $[+0.110, +0.217]$, all three decided. The prediction pools
each pair's two questions' own held-out readings, weighted by their need counts, and no departure from
it is detected: the composed gain minus the predicted one is $+0.019$ $[-0.025, +0.064]$ for coverage
and $+0.018$ $[-0.037, +0.077]$ for depth-weighted recall, over the $43$ pairs whose two questions both
have held-out runs. At each policy's own stop the composed gain exceeds the prediction, by $+0.111$
$[+0.059, +0.166]$ in coverage, a reading about stopping rather than equal spend. The pairing is ours,
so these are composed natural tasks and not a natural benchmark, and answer correctness is not read,
because the composed answer has no aliases.
A comparison against GPT-OSS-120B, prompted, on the same tasks was declared after the result above had
been read and before its own first unit, so it is exploratory. At equal spend the trained questioner covers more of the required
evidence than GPT-OSS-120B, by $+0.083$ $[+0.044, +0.124]$, and weights more of it toward deep needs, by
$+0.082$ $[+0.028, +0.137]$. Both are decided. Its lead in breadth, $+0.073$ $[-0.034, +0.182]$, is not.

\textbf{Stopping, read per decision point.} An episode of $n$ questions offers $n+1$ decisions, and a
run-level summary keeps only the last, so a run that asked six questions after it was done and then
stopped would score a perfect one. A point is done when the coverage the policy held then was
complete, and one whose coverage cannot be read is skipped. Every point before the last is an ask,
and the last is a stop only when the policy chose to stop. A state the harness halted at its ceiling
or turn limit was never offered to the policy, so it carries no decision and is counted apart as a
forced stop. Each rate is reported with its denominator, a property of the policy, beside the
questions bought at points already done.

\textbf{Quantities that are structurally determined, and quantities deleted.} A task's total facet
count, fixed by its graph, and stop undershoot, zero by algebra, cannot take another value and so are
never cited as nulls. Five candidate metrics were deleted:
anticipation depth (non-monotone), proactive precision as first stated (circular), the probability of
asking at least one user-directed question (a property of the ordering), stop regret in quality units
(it charges a perfect stopper) and future conversation coverage (it scores a miss on exactly the
success case).

\textbf{Suites, splits and eligibility.} FRAMES, $824$ tasks, ships no need graph, so coverage, depth
and breadth are undefined there rather than zero. The tool-use domains are never mined, and runs forked
from a dialogue prefix another system generated are flagged exploratory
(Appendix~\ref{app:transfer}). The partition is a property of the task
identifier: a bucket hashed from the suite and the template identifier where one exists, else the
task identifier, with buckets $0$ to $59$ training, $60$ to $74$ development and $75$ to $99$ held
out, so two instantiations of one template never straddle the wall. A row reaches a
question-answering table only if its split is held out, it is not flagged exploratory, and the run
reconciles on tokens and on documents. Each suite carries $200$ distinct eligible held-out tasks, out
of held-out corpora of $214$, $581$ and $3{,}080$. MuSiQue's held-out partition counts $212$ tasks when
its adapter is constructed directly and $214$ over run directories. All $200$ reported MuSiQue tasks are
held out under the split function itself, so the difference cannot move a task into the reported population.

\textbf{Arms and the estimator.} Every arm shares a ceiling of eight questions and a limit of sixteen
turns, and the number of passages and a hash of the pool are inside a run's identity, so a different
index is a different run rather than a confound. Every difference is paired on suite, task and seed,
and seeds and any other replicate of a task, such as a prompt variant, are averaged into the task
before resampling, so a count of pairs is never an $n$. Intervals are task-clustered bias-corrected
and accelerated bootstraps printed at ten thousand resamples, and a cell is decided only when its
intervals at fifty thousand resamples under three bootstrap seeds all exclude zero
(Section~\ref{sec:setup}). No endpoint of the preregistered stage names a trained arm, so every trained-arm contrast on the
question-answering suites is exploratory and is read from its estimate and interval. The prompt search's probability values are the
measurement rather than the evidence, since a searched prompt's apparent gain on the slice that
selected it did not hold on fresh tasks (Appendix~\ref{app:gepa}). Every number belongs to one
scorer version, which composes the metric register, graph set, matcher and judge pins, and
comparisons stay within one.

\textbf{What is new, and what is borrowed.} The estimator families are ordinary: coverage is set
recall, depth-weighted recall a discounted-gain recall \citep{jarvelin2002cumulated} weighted by depth
instead of rank, breadth an aspect count \citep{clarke2008novelty} read from the graph, question
diversity distinct-$n$ \citep{li2016diversity}, cost per need a cost-normalized effectiveness
\citep{erol2025costofpass}, and the stopping pair's nearest relative abstention
\citep{kirichenko2025abstentionbench,zhai2026abstainr1}. What we claim is the substrate, a
required-evidence graph whose edges are prerequisites, which makes $d_v$ and $r_v$ definable, and
three readings on it: the out-of-order rate and promptness, which we have not found elsewhere as
transcript-level readings, and the stopping pair read at every decision point, a change of unit rather
than of formula. The paper as a whole combines this content axis with candidates sampled at one state
and ordered by consequence, comparison at equal retrieval spend (Appendix~\ref{app:matchedcost}), and
zero-shot transfer to a tool-use benchmark with a simulated customer
\citep{yao2024taubench,barres2025tau2}. Two of these parts are stated more specifically: one model running
the protocol is equivalent to the two-model split at the shared cap and separated from it at equal
spend (Appendix~\ref{app:controls}), and on the tool-use benchmark the trained questioner raises task
success, with follow-up turns reported in Appendix~\ref{app:transfer}.

\textbf{Evaluation principles.} A move is credited by its consequence against the graph, never by how
it reads, the role grounding plays on the initiative axis \citep{bui2026proactivity}, and the two
axes are never pooled. When only a person can resolve a need, the
joint claim is two numbers, user turns and success, never one ratio \citep{yao2026bao}. Interruption
cost and adaptation after feedback stay on the initiative axis \citep{bui2026proactivity}, since this
questioner interrupts no one.
\FloatBarrier
\section{One task, three policies}\label{app:example}

This section supports Section~\ref{sec:axes}, and it illustrates the axis rather than measuring it: the aggregate reading is Table~\ref{tab:heldout}.

Figure~\ref{fig:example} shows one held-out task in full, with the questions each policy
asked in the order it asked them, and the required need each question resolved. The task's
third need is the clearest case of the vertical axis this paper can show: the film the
question asks about is named in it only by the start of its title, and where it is set cannot be
known until the director and then that director's birthplace are. The no-question baseline reports
that the evidence does not settle the question, which is true of the evidence it gathered. The
same model, prompted, resolves the first two needs, then asks for the birthplace it has already been given
four more times, and stops one need short after six questions. Both training seeds of the trained questioner
reach the third need first, on their third question, by searching for a work whose title begins with
the words the task supplies, before either prerequisite is in hand, one of them while naming the
wrong director. The second seed then resolves both prerequisites and stops after five questions with
every required need resolved, and the first resolves one of them and is stopped by the ceiling of
eight. That is out of order, but of the kind a question aimed at the need itself produces. Most of the
excess out-of-order rate Table~\ref{tab:heldout} reports on this suite is the other kind, evidence a question aimed
elsewhere brought back (Appendix~\ref{app:heldout}), so the example is not its typical source. It is an
evidence outcome, not an
answer outcome: the strict answer match is zero for all four runs, so the figure reports resolved
needs and not a correct answer.

\begin{figure}[p]
\centering
\scalebox{0.93}{\definecolor{pxhit}{HTML}{0B7A5D}
\definecolor{pxmiss}{HTML}{9A9A9A}
\definecolor{pxrule}{HTML}{C8C8C8}
\definecolor{pxband}{HTML}{F2F4F5}
\begin{tikzpicture}[
  font=\small,
  hit/.style={circle, fill=pxhit, text=white, inner sep=0pt, minimum size=3.6mm, font=\scriptsize\bfseries},
  miss/.style={circle, draw=pxmiss, text=pxmiss, inner sep=0pt, minimum size=3.6mm, font=\scriptsize},
  need/.style={draw=pxrule, fill=pxband, rounded corners=2pt, align=center, inner sep=3pt, font=\scriptsize, text width=2.7cm},
  panel/.style={draw=pxrule, rounded corners=2pt, line width=0.4pt},
  lbl/.style={font=\scriptsize, inner sep=1pt},
  arr/.style={-{Latex[length=1.5mm]}, draw=black!55},
  row/.style={anchor=north west, text width=12.45cm, align=left, inner ysep=1pt},
]
\coordinate (pxorigin) at (0,0);
\coordinate (pxright) at (13.60,0);
\coordinate (pxrulend) at (13.38,0);
\node[anchor=north west, font=\small\bfseries] at (0,0.00) {(a) What the task requires};
\node[anchor=north west, text width=13.40cm, align=left, font=\scriptsize] at (0,-0.50) {``When does Meet Me in the birthplace of Gracie's director take place?'' The film is named only by the start of its title, and where it is set cannot be known until the first two needs are resolved.};
\node[need, anchor=north west, text width=2.60cm, minimum height=0.92cm] (n0) at (0.00,-1.36) {who directed \emph{Gracie}};
\node[hit, anchor=center] at (0.00,-1.36) {0};
\node[need, anchor=north west, text width=2.60cm, minimum height=0.92cm] (n1) at (5.35,-1.36) {that director's\\birthplace};
\node[hit, anchor=center] at (5.35,-1.36) {1};
\draw[arr] (2.90,-1.82) -- (5.29,-1.82);
\node[lbl, anchor=south, align=center, text=black!65, fill=white, inner sep=1.5pt] at (4.12,-1.78) {Davis\\Guggenheim};
\node[need, anchor=north west, text width=2.60cm, minimum height=0.92cm] (n2) at (10.70,-1.36) {when the film set\\there takes place};
\node[hit, anchor=center] at (10.70,-1.36) {2};
\draw[arr] (8.25,-1.82) -- (10.64,-1.82);
\node[lbl, anchor=south, align=center, text=black!65, fill=white, inner sep=1.5pt] at (9.47,-1.78) {St.\,Louis};
\node[anchor=north west, font=\scriptsize, text=black!60] at (0,-2.32) {badge: the depth of the need};
\node[anchor=north west, font=\small\bfseries] at (0,-2.86) {(b) What each policy asked, and what it answered};
\coordinate (px0l) at (0,-3.48);
\coordinate (px0r) at (px0l -| pxright);
\node[anchor=north west, font=\small\bfseries] at ([xshift=0.22cm,yshift=-0.10cm]px0l) {No questions asked};
\node[anchor=north east, font=\scriptsize] at ([xshift=-0.22cm,yshift=-0.12cm]px0r) {0 of 3 needs \quad \textcolor{pxmiss}{needs unresolved}};
\node[row] (px0q0) at ([xshift=0.75cm,yshift=-0.58cm]px0l) {\strut \textcolor{pxmiss}{\itshape asks nothing}\strut};
\node[miss, anchor=center] at ([xshift=-0.39cm,yshift=0.10cm]px0q0.base west) {--};
\coordinate (px0s) at ([yshift=-0.04cm]px0q0.south west);
\draw[draw=pxrule] (px0s) -- (px0s -| pxrulend);
\node[anchor=north west, text width=12.45cm, align=left, font=\scriptsize] (px0a) at ([yshift=-0.04cm]px0s) {\textit{Answer.} The evidence does not specify when ``Meet Me in the birthplace of Gracie's director'' takes place. The date is unsupported by the provided evidence.};
\coordinate (px0bb) at (px0a.south -| pxorigin);
\coordinate (px0b) at ([yshift=-0.06cm]px0bb);
\node[panel, fit=(px0l)(px0r)(px0b), inner sep=0pt] {};
\coordinate (px1t) at (px0b -| pxorigin);
\coordinate (px1l) at ([yshift=-0.26cm]px1t);
\coordinate (px1r) at (px1l -| pxright);
\node[anchor=north west, font=\small\bfseries] at ([xshift=0.22cm,yshift=-0.10cm]px1l) {The same model, prompted};
\node[anchor=north east, font=\scriptsize] at ([xshift=-0.22cm,yshift=-0.12cm]px1r) {2 of 3 needs \quad \textcolor{pxmiss}{needs unresolved}};
\node[row] (px1q0) at ([xshift=0.75cm,yshift=-0.58cm]px1l) {\strut ``Who is Gracie's director?''\strut};
\node[hit, anchor=center] at ([xshift=-0.39cm,yshift=0.10cm]px1q0.base west) {0};
\node[row] (px1q1) at (px1q0.south west) {\strut ``Where was Davis Guggenheim's birthplace located?''\strut};
\node[hit, anchor=center] at ([xshift=-0.39cm,yshift=0.10cm]px1q1.base west) {1};
\node[row] (px1q2) at (px1q1.south west) {\strut ``What is the birthplace of Davis Guggenheim?'' \textcolor{pxmiss}{\scriptsize (asked 4 times)}\strut};
\node[miss, anchor=center] at ([xshift=-0.39cm,yshift=0.10cm]px1q2.base west) {--};
\coordinate (px1s) at ([yshift=-0.04cm]px1q2.south west);
\draw[draw=pxrule] (px1s) -- (px1s -| pxrulend);
\node[anchor=north west, text width=12.45cm, align=left, font=\scriptsize] (px1a) at ([yshift=-0.04cm]px1s) {\textit{Answer.} The time cannot be determined from the provided evidence; the evidence lacks information about when ``Meet Me'' takes place.};
\coordinate (px1bb) at (px1a.south -| pxorigin);
\coordinate (px1b) at ([yshift=-0.06cm]px1bb);
\node[panel, fit=(px1l)(px1r)(px1b), inner sep=0pt] {};
\coordinate (px2t) at (px1b -| pxorigin);
\coordinate (px2l) at ([yshift=-0.26cm]px2t);
\coordinate (px2r) at (px2l -| pxright);
\node[anchor=north west, font=\small\bfseries] at ([xshift=0.22cm,yshift=-0.10cm]px2l) {The trained questioner, training seed 1};
\node[anchor=north east, font=\scriptsize] at ([xshift=-0.22cm,yshift=-0.12cm]px2r) {2 of 3 needs \quad \textcolor{pxmiss}{needs unresolved}};
\node[row] (px2q0) at ([xshift=0.75cm,yshift=-0.58cm]px2l) {\strut ``Who is the director of the film or work titled `Gracie' and what is the director's birthplace?''\strut};
\node[miss, anchor=center] at ([xshift=-0.39cm,yshift=0.10cm]px2q0.base west) {--};
\node[row] (px2q1) at (px2q0.south west) {\strut ``What is the birthplace (city or location) of the director of the work titled `Gracie'?''\strut};
\node[miss, anchor=center] at ([xshift=-0.39cm,yshift=0.10cm]px2q1.base west) {--};
\node[row] (px2q2) at (px2q1.south west) {\strut ``What is the setting (city, town, or location) where the film or work titled `Meet Me in' is set, and in what time period or year does it take place?''\strut};
\node[hit, anchor=center] at ([xshift=-0.39cm,yshift=0.10cm]px2q2.base west) {2};
\node[row] (px2q3) at (px2q2.south west) {\strut ``Which paragraph mentions the director of the work titled `Gracie' and provides the director's birthplace (city or location)?''\strut};
\node[miss, anchor=center] at ([xshift=-0.39cm,yshift=0.10cm]px2q3.base west) {--};
\node[row] (px2q4) at (px2q3.south west) {\strut ``Which paragraph mentions the director of the work titled `Gracie' and also states the city or location where that director was born?''\strut};
\node[hit, anchor=center] at ([xshift=-0.39cm,yshift=0.10cm]px2q4.base west) {1};
\node[row] (px2q5) at (px2q4.south west) {\strut \textcolor{pxmiss}{\itshape three further rewordings of the question above, none resolving a need}\strut};
\node[miss, anchor=center] at ([xshift=-0.39cm,yshift=0.10cm]px2q5.base west) {--};
\node[row] (px2q6) at (px2q5.south west) {\strut \textcolor{pxmiss}{\scriptsize\itshape stopped by the budget of eight}\strut};
\coordinate (px2s) at ([yshift=-0.04cm]px2q6.south west);
\draw[draw=pxrule] (px2s) -- (px2s -| pxrulend);
\node[anchor=north west, text width=12.45cm, align=left, font=\scriptsize] (px2a) at ([yshift=-0.04cm]px2s) {\textit{Answer.} Summer 1903 through spring 1904.};
\coordinate (px2bb) at (px2a.south -| pxorigin);
\coordinate (px2b) at ([yshift=-0.06cm]px2bb);
\node[panel, fit=(px2l)(px2r)(px2b), inner sep=0pt] {};
\coordinate (px3t) at (px2b -| pxorigin);
\coordinate (px3l) at ([yshift=-0.26cm]px3t);
\coordinate (px3r) at (px3l -| pxright);
\node[anchor=north west, font=\small\bfseries] at ([xshift=0.22cm,yshift=-0.10cm]px3l) {The trained questioner, training seed 2};
\node[anchor=north east, font=\scriptsize] at ([xshift=-0.22cm,yshift=-0.12cm]px3r) {3 of 3 needs \quad \textcolor{pxhit}{every need resolved}};
\node[row] (px3q0) at ([xshift=0.75cm,yshift=-0.58cm]px3l) {\strut ``Who is the director of the film or work titled `Gracie' and what is the birthplace of that director?''\strut};
\node[miss, anchor=center] at ([xshift=-0.39cm,yshift=0.10cm]px3q0.base west) {--};
\node[row] (px3q1) at (px3q0.south west) {\strut ``What is the birthplace (city and country) of film director Nelson Cragg, the director of photography for the series Homeland?''\strut};
\node[miss, anchor=center] at ([xshift=-0.39cm,yshift=0.10cm]px3q1.base west) {--};
\node[row] (px3q2) at (px3q1.south west) {\strut ``What is the birthplace (city and country) of film director Nelson Cragg, the director of photography for the series Homeland, and does any paragraph describe a work titled `Meet Me in' that takes place in that city?''\strut};
\node[hit, anchor=center] at ([xshift=-0.39cm,yshift=0.10cm]px3q2.base west) {2};
\node[row] (px3q3) at (px3q2.south west) {\strut ``Which paragraph describes a film, book, or other work titled `Gracie' and names its director?''\strut};
\node[hit, anchor=center] at ([xshift=-0.39cm,yshift=0.10cm]px3q3.base west) {0};
\node[row] (px3q4) at (px3q3.south west) {\strut ``What is the birthplace (city and country) of film director Davis Guggenheim, the director of the film Gracie?''\strut};
\node[hit, anchor=center] at ([xshift=-0.39cm,yshift=0.10cm]px3q4.base west) {1};
\coordinate (px3s) at ([yshift=-0.04cm]px3q4.south west);
\draw[draw=pxrule] (px3s) -- (px3s -| pxrulend);
\node[anchor=north west, text width=12.45cm, align=left, font=\scriptsize] (px3a) at ([yshift=-0.04cm]px3s) {\textit{Answer.} Summer 1903 -- spring 1904.};
\coordinate (px3bb) at (px3a.south -| pxorigin);
\coordinate (px3b) at ([yshift=-0.06cm]px3bb);
\node[panel, fit=(px3l)(px3r)(px3b), inner sep=0pt] {};
\end{tikzpicture}}
\caption{One held-out MuSiQue task, and what each policy asked. A filled badge on a question gives the
depth of the required need that question resolved, and a hollow badge marks a question that
resolved none. \textbf{This is an illustration and not evidence.} The trained questioner is shown at
both of its training seeds, and the task was chosen by a rule fixed before any of its transcripts was
read: among the $175$ held-out tasks of this suite where every policy has a completed run at the same
rollout seed, take those where each training seed resolves a need at depth two or more that the
same model, prompted, does not, and select the first by task identifier. Sixteen tasks satisfy that rule and
three satisfy it with the two policies exchanged, so the direction shown here is the more common one
and not the only one. The chain's labels are glossed from the benchmark's own decomposition forms,
which the released record carries verbatim.}
\label{fig:example}
\end{figure}
\FloatBarrier
\section{Comparison at equal retrieval spend}\label{app:matchedcost}

\begin{figure}[t]
\centering
\includegraphics[width=\linewidth]{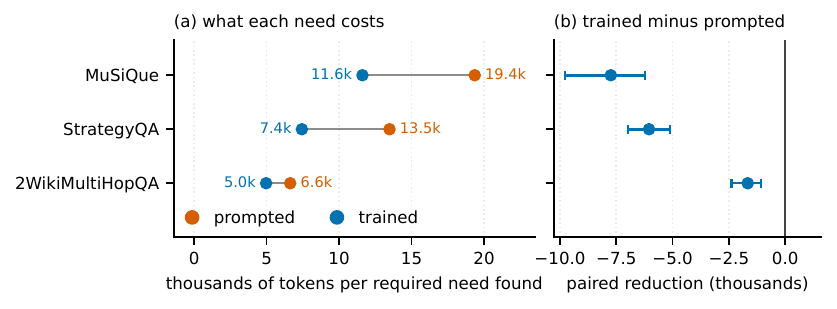}
\caption{Efficiency with no matching rule. (a) Tokens spent per required need found, lower being better, as paired means over the tasks both arms completed.
(b) The paired reductions, every interval excluding zero. The trained questioner's two training seeds
are averaged within each task, against the same model, prompted.}
\label{fig:efficiency}
\end{figure}

This appendix supports the equal-spend rule of Table~\ref{tab:heldout}, the efficiency reading of
Figure~\ref{fig:efficiency}, the claim of Section~\ref{sec:length} that the gain is not bought
with longer questions, and the structured-retrieval comparison of Figure~\ref{fig:comparators}b.

\textbf{A prefix of a long run is a short run.} The policy that chooses questions is a function of
the dialogue state alone, which holds no ledger of spend, no remaining allowance and no cap: the usage
stamped on past turns says what has been spent, not where the ceiling is, and truncation is applied
from outside and never announced. The question it asks fifth is therefore the one it would have asked
had its cap been five, so coverage after $k$ questions of a long run is the terminal coverage of a
true $k$-question run. The equal-spend rule rests on this: the prefix read from the arm that asked
more is a run it would in fact have produced. A budget-aware policy would lose the property twice
over, spending its last question differently for knowing it was the last and stopping as an announced cap approached, so the
stopping measurement would report the experiment's own parameter back to it.

\textbf{The assumption, tested.} To build the policy the argument rules out, GPT-OSS-120B, prompted,
was given one added prompt block stating in words how many questions it may ask, and run against its
unmodified self at a retrieval cap of four on all three suites, $200$ tasks at two seeds each. Its
required-evidence coverage rises on none of them, with paired differences of $-0.0244$ $[-0.0496,
+0.0004]$ on MuSiQue, $+0.0033$ $[-0.0162, +0.0231]$ on StrategyQA and $-0.0056$ $[-0.0200,
+0.0075]$ on 2WikiMultiHopQA, and the informed comparator asks slightly fewer questions rather than
better early ones, $3.07$ against $3.24$, $3.28$ against $3.41$ and $2.38$ against $2.42$. This
exploratory control, not a sealed endpoint, tests whether a comparator told its allowance asks better
early questions, not whether a truncated trajectory equals a true short run, since both arms here
were given a true cap.

\textbf{What is read at the shared cap alone.} A shared cap charges the arm that stops earlier for
the calls it declined and credits nothing for the calls it saved, so a reading at the shared cap answers a
different question, which policy does better when each decides when to stop (Table~\ref{tab:regimes}),
and is not a second reading of evidence per question.
Stop overshoot has no equal-spend form, because the rule sets the comparator's stopping index to the
trained arm's, the quantity the metric measures. Nor does answer quality, since a prefix-length answer
would need a drafting call that was not taken. The selection gate records coverage at depth two or
more at the shared cap only, since rebuilding it at a prefix needs the gold map from evidence to
depth, which the gate may not read.

\textbf{The currency is retrieval calls, and one question is one call.} On these suites question
count and retrieval calls are the same quantity on every run, so equal questions is equal retrieval
spend by identity and not by approximation. The identity does not travel: in the tool-use benchmark
of Appendix~\ref{app:transfer} a question costs a variable number of tool calls, so there questions
and tool calls are two currencies, each enforced separately. A third currency, unique documents
reaching the drafter, is neither matched nor reported, and its unmatched level points the other way:
on held-out MuSiQue the trained questioner asks $1.41$ times fewer questions than the same model,
prompted, but sees $2.6\%$ more documents, since the comparator's extra questions retrieve
heavily overlapping documents.

\textbf{A reading that needs no matching rule.} Tokens spent per required need the run actually found
put the cost in the denominator, so no prefix is read, no pairing convention enters, and the currency
means the same for a locally served policy and a hosted one. On paired held-out tasks the
trained questioner's reductions are $-7{,}744$ $[-9{,}786, -6{,}222]$, $-6{,}046$ $[-6{,}992,
-5{,}135]$ and $-1{,}663$ $[-2{,}386, -1{,}065]$ on MuSiQue, StrategyQA and 2WikiMultiHopQA, from
$19{,}354$ tokens per need found to $11{,}610$ on MuSiQue (Figure~\ref{fig:efficiency}).

\textbf{Question length, and the budget that controls for it.} A longer question costs the same
single retrieval call, so no call-denominated comparison can test whether the gain was bought with
words rather than with better questions. On development ($132$ MuSiQue and $333$ StrategyQA
tasks) the trained questioner's questions average $17.5$ and $22.6$ words against
$13.1$ and $13.0$ for the same model, prompted, and $21.7$ and $29.2$ question tokens against $16.0$ and $15.6$. It
fails the one-sided not-longer rule, whose declared margin is $0.1$, on both suites and in both units,
with pooled relative lower bounds of $+0.304$ and $+0.320$ on MuSiQue and $+0.705$ and $+0.829$ on
StrategyQA, in words and in tokens, and each seed fails alone.

The control therefore charges the comparator the trained questioner's own question budget in
question tokens, the text the drafter receives as the model counts it, with question words as the
check. The comparator is granted complete questions in the order it asked them until the next would overrun the allowance, so its realized spend sits below the
allowance. The equal-call reading it is compared with is the selection gate's form of the rule: the
trained questioner at its natural stop, the comparator at its own prefix of the same length or its
own end, whichever comes first. That matches spend in one direction only: where the comparator stops
first, the trained questioner keeps its extra calls. This appendix calls it gate-matched spend and keeps
equal spend for both arms read at the lower count.

\begin{table}[h]
\centering
\caption{Coverage when the comparator is charged the questioner's own question budget, on development. The trained questioner averages seeds 1 and 2 within each task. Intervals are the task-clustered bootstrap at $1{,}000$ resamples.}
\label{tab:app-lengthbudget}
\resizebox{\linewidth}{!}{\begin{footnotesize}\begin{tabular}{llrrrrr}
\toprule
suite & unit & allowance & comparator & comparator & $\Delta$ coverage & 95\% interval \\
 & & granted & spend & questions & under budget & \\
\midrule
MuSiQue            & question tokens & 91.8 & 65.5 & 4.261 & $+0.1487$ & $[+0.1130, +0.1949]$ \\
MuSiQue            & question words  & 74.3 & 53.3 & 4.250 & $+0.1534$ & $[+0.1174, +0.2019]$ \\
\addlinespace
StrategyQA         & question tokens & 62.2 & 42.0 & 2.974 & $+0.0466$ & $[+0.0255, +0.0692]$ \\
StrategyQA         & question words  & 48.2 & 33.2 & 2.881 & $+0.0515$ & $[+0.0295, +0.0766]$ \\
\bottomrule
\end{tabular}\end{footnotesize}}
\end{table}

\noindent Charged that budget, the comparator does not catch up, and all four intervals exclude zero.
At equal retrieval calls it was charged $3.981$ and $2.080$ questions and trailed by $+0.1556$ and
$+0.0799$ on MuSiQue and StrategyQA. On StrategyQA the budget buys it $43\%$ more asking and the
difference falls to $+0.0466$, and on MuSiQue it moves to $+0.1487$. The allowance exceeds everything
the comparator asks within eight calls on $41\%$ to $45\%$ of MuSiQue tasks, so the comparator was
free to spend the verbal budget on more questions and could not convert it into evidence: the gain is
in what was asked, not in how much was said. The instrument is not generous to every arm. The arm
trained on stop contrasts alone, which asks the most questions of any trained arm, fails both tests
on the same tasks: its differences at matched retrieval calls, $-0.0051$ and $-0.0169$, and all four under its own question budget have
intervals containing zero.

\textbf{A structured retrieval method on three models (Figure~\ref{fig:comparators}b).} PAR$^2$-RAG
\citep{li2026par2rag}, reimplemented without training in the same harness (Section~\ref{sec:related}),
never beats plain prompting of its own model at equal retrieval spend, on Qwen3-8B, GPT-OSS-120B or Claude Opus 5. On the larger two
it is lower, by $-0.037$ $[-0.070, -0.005]$ on MuSiQue for GPT-OSS-120B and by $-0.074$ and $-0.017$
on MuSiQue and StrategyQA for Claude Opus 5. The trained questioner leads the reimplementation by $+0.100$ and
$+0.060$ on Qwen3-8B, its own base weights, by $+0.124$ and $+0.055$ on GPT-OSS-120B, and on Claude
Opus 5 by $+0.046$ on MuSiQue. The one comparator ahead of it is Claude Opus 5 prompted plainly, on
MuSiQue, where the trained questioner minus it is $-0.029$ $[-0.055, -0.003]$, and on MuSiQue
Claude Opus 5 asks fewer questions, $3.5$ against the trained questioner's $4.2$.

\textbf{On FRAMES.} FRAMES has no need graph, so the third panel reads answer correctness at each
policy's own stop under a ceiling of eight calls, over $824$ tasks, and is not a comparison at equal
spend. No difference between the trained questioner and a comparator prompted plainly is decided, from
$-0.010$ $[-0.039, +0.018]$ against GPT-OSS-120B to $+0.014$ $[-0.016, +0.044]$ against the same model,
and none against the reimplementation on Qwen3-8B or GPT-OSS-120B. Claude Opus 5 running PAR$^2$-RAG is ahead:
the trained questioner minus it is $-0.061$ $[-0.093, -0.032]$, decided, with $0.237$ of tasks answered
correctly against $0.299$, and on FRAMES the trained questioner uses $3.6$ retrieval calls against
its $4.5$, so Claude Opus 5 running PAR$^2$-RAG spends more to lead.

\FloatBarrier
\section{Corpus, export rules and training recipes}\label{app:training}

This section gives the corpus, export rules and recipes behind the trained questioner of
Section~\ref{sec:questioner} and the supervision comparison of Table~\ref{tab:mechanism}, and reads
every arm on the held-out split.

\textbf{What the corpus holds.} A decision is a state, rendered exactly as the policy sees it at
inference, an action taken there, and a label derived from what that action caused. At a chosen state of
a recorded run we sample eight alternative questions at temperature one, with no nucleus or top-$k$
truncation, and roll each to completion. The eight share the state byte for byte, so whatever value it
carries cancels within a pair and the preference needs neither a baseline nor a critic. Each is labeled
with whether its episode answered the task, how many turns it still needed to reach complete evidence,
and whether it resolved a need as soon as that need became askable. The widest export, with no
filter on which arm produced a run, holds $44{,}624$ imitation rows and $50{,}730$ preference pairs over
$32{,}958$ states and $1{,}918$ tasks. Of its pairs, $16{,}089$ set one question against another,
$9{,}668$ of them at a state where evidence has already arrived. The rest set a question against stopping, $1{,}958$ where a candidate actually stopped and
$32{,}683$ derived from the need graph at complete states where no candidate stopped, kept separately
selectable because a gold-derived stop is not the same evidence as one a policy took.

\textbf{The export rules.} The export reads the need graphs to value every decision point, so it runs on
the gold side and the trainer never does. It re-renders the prompt the policy saw from the recorded
artifacts and refuses a run whose evidence hash does not re-derive or whose draft is not recoverable.
Every prompt that shows evidence, in every arm and in the training data alike, cuts each retrieved
paragraph at its first $1{,}200$ characters, which $3.5\%$, $6.1\%$ and $5.2\%$ of the paragraphs in
the three corpora exceed. The acceptance floor is specified as a percentile of pilot values plus a term
in the judge's deviation, but the export ran at its command-line defaults, a floor of $0.05$ and a judge
deviation of zero, so the rule that ran was a flat threshold at $0.05$ on the consequence score.

\textbf{The imitation signal, and the refusal.} The imitation target is a stop where the required
evidence was already in hand before the decision, and otherwise the best sampled question if it clears
the floor. Where nothing clears the floor, no row is emitted. That refusal discarded $5{,}030$
unfinished states from the fixed-cohort export the imitation reference trained on and $24{,}538$ from
the pooled export, which keeps only runs of the same model, prompted, and applies no cohort filter. The
pooled export is not the widest export above, which admits every arm.

\textbf{The preference signal.} The pairs the trained questioner's preference stage trains on are of three kinds. A question pair sets two sampled questions
against each other, and a recorded stop contrast sets a question against a stop the policy actually
took. A synthesized stop contrast, emitted at a state the graph marks unfinished where some candidate
asked and none stopped, chooses the best available question over stopping, which restores as contrasts
the states the imitation refusal drops. It is never emitted at a state of unknown completion, and a
state where a candidate actually stopped keeps its recorded pair instead. Sampled questions are ordered
by outcome, first by whether the episode answered the task (Section~\ref{sec:questioner}). A key ranking
anticipation first is only a flag, because under it the finding that anticipation emerges from ranking
on outcome would hold by construction.

\begin{table}[ht]
\centering
\caption{The arms. The reference is an imitation adapter fitted to the $14{,}302$ recorded decisions
of the same model, prompted, in the fixed-cohort export, two epochs at learning rate $10^{-4}$, every
other arm varies one thing against it, and the trained questioner is the last row. Its combined set holds $11{,}305$ question pairs,
$1{,}249$ recorded stop contrasts and $18{,}919$ synthesized ones, $31{,}473$ of the $36{,}349$ pairs
in its preference file, the $4{,}876$ decided by a finished state being left out of training. By
source they are $19{,}448$ MuSiQue, $9{,}786$ StrategyQA and $2{,}232$ 2WikiMultiHopQA pairs and $7$
synthesized stop contrasts from a small synthetic fixture. The
stop-contrast-only set is those $18{,}919$ synthesized contrasts alone, and the question-pair arm's
$17{,}430$-pair export is the fixed-cohort preference export. Every adapter uses rank $32$,
scaling $64$, dropout $0.05$ and no bias term, over the same seven projection modules. The preference
arms train for one epoch at learning rate $5\times10^{-6}$ with a divergence weight of $0.1$.}
\label{tab:app-train-recipes}
\resizebox{\linewidth}{!}{\begin{tabular}{llll}
\toprule
arm & fitted from & training records & what it varies \\
\midrule
imitation reference        & base model          & $14{,}302$ decisions & nothing (the comparison point) \\
stop-weighted imitation    & base model          & $14{,}302$ decisions & stop rows down-weighted to $0.1$ \\
question pairs             & imitation reference & question pairs of a $17{,}430$-pair export & preference on question pairs \\
stop contrasts only        & imitation reference & $18{,}919$ pairs, one kind & preference on stop contrasts only \\
both kinds, from reference & imitation reference & $31{,}473$ pairs, three kinds & preference on both kinds at once \\
both kinds, from pairs     & question pairs      & $31{,}473$ pairs, three kinds & the same set, from a trained start \\
\bottomrule
\end{tabular}}
\end{table}

\textbf{How the arms were compared on development.} Each arm of Table~\ref{tab:app-train-recipes} is
read as a paired required-evidence coverage difference against the same model, prompted, at
gate-matched spend, at its own stopping point against the comparator's prefix of that length
(Appendix~\ref{app:matchedcost}). Intervals are task-clustered bias-corrected bootstraps at $1{,}000$
resamples over $132$ MuSiQue and $333$ StrategyQA development tasks, so this is a selection record
rather than a confirmatory result. Of the arms whose development cells are printed
(Tables~\ref{tab:mechanism} and~\ref{tab:app-stop-composition}), every one but the arm trained on stop
contrasts alone gains coverage on both suites with an interval excluding zero. What each signal moves,
in coverage, persistence and stopping when done, is read in Appendix~\ref{app:stopping}.

\textbf{Ordering rule and selection.} Three arms that vary only the rule ordering the sampled questions
all gain coverage, and against one another only the consequence ordering separates from an ordering by
whether a question reaches an unstated need, on StrategyQA, at $+0.0139$ $[+0.0025, +0.0293]$, a
separation that does not survive a correction for all six contrasts, so which
ordering to prefer is exploratory. Of the twenty-three development checkpoints, seventeen gain coverage
at gate-matched spend on both suites with an interval excluding zero, and none passes the pre-committed
selection rule as written. The checkpoint carried forward was chosen on another pre-committed ground,
that an arm keeps both stopping cells and gains coverage, read on the first registered training run of
the recipe (Appendix~\ref{app:seedzero}). The trained questioner's two training seeds gain coverage at
gate-matched spend on both suites, each interval excluding zero, but do not keep the stopping cell
(Appendix~\ref{app:stopping}).

\textbf{Question length.} The trained questioner fails the gate's length test on both suites
(Table~\ref{tab:app-stop-length}), and Appendix~\ref{app:matchedcost} reads coverage with the
comparator charged the trained questioner's own question budget (Section~\ref{sec:length}).

\begin{table}[ht]
\caption{The same arms on the held-out split, each against the same model, prompted, at
gate-matched spend (Appendix~\ref{app:matchedcost}), $200$ tasks per suite under one contrast function. The last
row, both pair kinds fitted to the question-pair arm, has the largest point estimate on every suite,
about a hundredth above the question-pair arm it starts from, a point ordering and not a tested
difference. The two question-pair rows start from different imitation checkpoints, the first from the
reference row above and the second from the one fitted to the pooled export, the question-pair row of
Table~\ref{tab:mechanism}. Intervals are bias-corrected bootstraps at $10{,}000$ resamples, with any
bound within $0.01$ of zero re-read at $50{,}000$ across three reseeds.}
\label{tab:app-heldout-methods}
\begin{center}\small
\begin{tabular}{@{}>{\raggedright\arraybackslash}p{4.2cm}@{\hspace{5pt}}c@{\hspace{5pt}}c@{\hspace{5pt}}c@{}}
\toprule
supervision & MuSiQue & StrategyQA & 2WikiMultiHopQA \\
\midrule
imitation reference & $+0.0858$ & $+0.0271$ & $+0.0587$ \\
\leavevmode{\scriptsize\color{black!60} filtered demonstrations} & {\scriptsize $[+0.0492, +0.1231]$} & {\scriptsize $[-0.0070, +0.0631]$} & {\scriptsize $[+0.0299, +0.0938]$} \\[5pt]
stop-weighted imitation & $+0.0746$ & $+0.0260$ & $+0.0594$ \\
\leavevmode{\scriptsize\color{black!60} the same, stop rows down-weighted} & {\scriptsize $[+0.0410, +0.1100]$} & {\scriptsize $[-0.0106, +0.0638]$} & {\scriptsize $[+0.0325, +0.0925]$} \\[5pt]
question pairs & $+0.1177$ & $+0.0693$ & $+0.0719$ \\
\leavevmode{\scriptsize\color{black!60} question pairs, from the reference} & {\scriptsize $[+0.0848, +0.1531]$} & {\scriptsize $[+0.0337, +0.1067]$} & {\scriptsize $[+0.0456, +0.1025]$} \\[5pt]
question pairs, pooled & $+0.1110$ & $+0.0657$ & $+0.0413$ \\
\leavevmode{\scriptsize\color{black!60} question pairs, from the pooled imitation checkpoint} & {\scriptsize $[+0.0797, +0.1448]$} & {\scriptsize $[+0.0319, +0.1008]$} & {\scriptsize $[+0.0125, +0.0725]$} \\[5pt]
stop contrasts only$^{\ddagger}$ & $+0.0177$ & $-0.0623$ & $+0.0175$ \\
\leavevmode{\scriptsize\color{black!60} stop contrasts, from the reference} & {\scriptsize $[-0.0142, +0.0496]$} & {\scriptsize $[-0.0961, -0.0304]$} & {\scriptsize $[-0.0075, +0.0469]$} \\[5pt]
both kinds, from reference & $+0.0800$ & $+0.0127$ & $+0.0625$ \\
\leavevmode{\scriptsize\color{black!60} both pair kinds, from the reference} & {\scriptsize $[+0.0446, +0.1165]$} & {\scriptsize $[-0.0237, +0.0503]$} & {\scriptsize $[+0.0344, +0.0975]$} \\[5pt]
both kinds, from pairs$^{\dagger}$ & $\mathbf{+0.1284}$ & $\mathbf{+0.0799}$ & $\mathbf{+0.0791}$ \\
\leavevmode{\scriptsize\color{black!60} both pair kinds, from the question-pair arm} & {\scriptsize $[+0.1021, +0.1593]$} & {\scriptsize $[+0.0511, +0.1116]$} & {\scriptsize $[+0.0519, +0.1113]$} \\[5pt]
\bottomrule
\multicolumn{4}{@{}p{13.6cm}@{}}{\scriptsize $\dagger$ the trained questioner, its two training seeds averaged within each task. $\ddagger$ this arm asks more questions than its comparator, so the rule every other row uses is not a matched reading for it, and its cells are the symmetric reading with both sides held to the lower question count. \textbf{Bold} marks the largest value in a column.} \\
\end{tabular}
\end{center}
\end{table}

\FloatBarrier
\section{Held-out results in full}\label{app:heldout}

This section carries the held-out cells behind the coverage, depth and ordering claims of
Table~\ref{tab:heldout}. No declared endpoint of the preregistered stage names the trained
questioner as a treatment, so as contrasts these readings are exploratory as a class, reported as
point estimates and intervals with no probability value.

\textbf{The differences behind Table~\ref{tab:heldout}.} Table~\ref{tab:heldout} prints levels, and
Table~\ref{tab:app-headline-diffs} prints each of its pairs as the trained questioner's difference from the
comparator, computed before rounding, with its interval.

\begin{table}[ht]
\centering
\caption{Table~\ref{tab:heldout} as differences: trained minus the named comparator over $200$ held-out
tasks per suite ($178$, $83$ and $113$ on the out-of-order row, where it is defined), with bias-corrected
intervals over tasks at $10{,}000$ resamples. $^{\circ}$ marks a cell not decided (Section~\ref{sec:setup})
and $\dagger$ a row where lower is better. Differences are in percentage points, except breadth (in lines)
and the deepest need resolved (in levels).}
\label{tab:app-headline-diffs}
\small
\begin{tabular}{@{}l@{\hspace{5pt}}c@{\hspace{5pt}}c@{\hspace{5pt}}c@{}}
\toprule
 & MuSiQue & StrategyQA & 2WikiMultiHopQA \\
\midrule
\multicolumn{4}{@{}l}{Equal spend: both arms at the lower of their two question counts} \\
required-evidence coverage & $+11.2$ {\scriptsize $[+8.7, +14.3]$} & $+7.0$ {\scriptsize $[+4.0, +10.3]$} & $+5.0$ {\scriptsize $[+1.7, +8.4]$} \\
breadth & $+0.120$ {\scriptsize $[+0.084, +0.165]$} & $+0.050$ {\scriptsize $[+0.008, +0.098]$} & $+0.070$ {\scriptsize $[+0.015, +0.128]$} \\
depth-weighted recall & $+12.5$ {\scriptsize $[+9.5, +16.1]$} & $+5.4$ {\scriptsize $[+2.6, +8.4]$} & $+5.5$ {\scriptsize $[+2.0, +9.3]$} \\
deepest need resolved & $+0.218$ {\scriptsize $[+0.151, +0.306]$} & $+0.098$ {\scriptsize $[+0.035, +0.176]$} & $+0.056$ {\scriptsize $[+0.001, +0.118]$}$^{\circ}$ \\
out-of-order rate$^{\dagger}$ & $+5.7$ {\scriptsize $[+0.7, +10.6]$} & $+0.5$ {\scriptsize $[-4.9, +5.3]$}$^{\circ}$ & $-1.2$ {\scriptsize $[-5.3, +1.9]$}$^{\circ}$ \\
coverage vs GPT-OSS-120B & $+7.0$ {\scriptsize $[+4.4, +9.6]$} & $+4.3$ {\scriptsize $[+1.5, +7.4]$} & $-3.5$ {\scriptsize $[-7.1, -0.3]$} \\
\midrule
\multicolumn{4}{@{}l}{Own stop: each arm stops itself, with at most eight calls} \\
required-evidence coverage & $+9.6$ {\scriptsize $[+7.1, +12.5]$} & $-1.2$ {\scriptsize $[-3.7, +1.4]$}$^{\circ}$ & $+1.5$ {\scriptsize $[-0.6, +4.0]$}$^{\circ}$ \\
coverage vs GPT-OSS-120B & $+10.1$ {\scriptsize $[+7.5, +13.0]$} & $-3.6$ {\scriptsize $[-5.8, -1.5]$} & $-0.7$ {\scriptsize $[-2.6, +0.8]$}$^{\circ}$ \\
answer token F1 & $+2.4$ {\scriptsize $[-0.6, +5.8]$}$^{\circ}$ & $-0.3$ {\scriptsize $[-4.3, +3.7]$}$^{\circ}$ & $+3.1$ {\scriptsize $[-0.5, +6.9]$}$^{\circ}$ \\
\bottomrule
\end{tabular}

\end{table}

\textbf{Each seed alone.} Table~\ref{tab:app-seeds} prints every Table~\ref{tab:heldout} cell for each
of the two training seeds (Section~\ref{sec:setup}) and for the two averaged within the task,
against one set of comparator runs under one rule. Each seed alone decides every coverage cell, and
the two agree in sign on every cell but the out-of-order rate on StrategyQA, where both span zero.

\begin{table}[ht]
\centering
\caption{Table~\ref{tab:heldout}'s cells per training seed, trained minus the same model, prompted, at
equal spend, $n=200$ tasks per suite on every row but the out-of-order one. The last column averages the two seeds within each task, and its intervals and
verdicts are in Table~\ref{tab:heldout}. The out-of-order rate is read only on the pairs where both
arms define it, so each seed's cell is over its own tasks and the averaged cell over every task where
either seed defines it, on MuSiQue $175$ and $174$ tasks against $178$. The averaged cell is therefore
not the midpoint of the two seeds' cells. $^{\dagger}$ Lower is better.}
\label{tab:app-seeds}
\small
\begin{tabular}{@{}llrrr@{}}
\toprule
 & suite & seed 1 & seed 2 & seeds 1+2 \\
\midrule
required-evidence coverage & MuSiQue & $+0.108$ & $+0.116$ & $+0.112$ \\
 & StrategyQA & $+0.064$ & $+0.076$ & $+0.070$ \\
 & 2WikiMultiHopQA & $+0.048$ & $+0.051$ & $+0.050$ \\
\addlinespace[2pt]
breadth & MuSiQue & $+0.120$ & $+0.120$ & $+0.120$ \\
 & StrategyQA & $+0.043$ & $+0.058$ & $+0.050$ \\
 & 2WikiMultiHopQA & $+0.065$ & $+0.075$ & $+0.070$ \\
\addlinespace[2pt]
depth-weighted recall & MuSiQue & $+0.122$ & $+0.129$ & $+0.125$ \\
 & StrategyQA & $+0.048$ & $+0.059$ & $+0.054$ \\
 & 2WikiMultiHopQA & $+0.052$ & $+0.058$ & $+0.055$ \\
\addlinespace[2pt]
deepest need resolved & MuSiQue & $+0.218$ & $+0.218$ & $+0.218$ \\
 & StrategyQA & $+0.090$ & $+0.105$ & $+0.098$ \\
 & 2WikiMultiHopQA & $+0.055$ & $+0.058$ & $+0.056$ \\
\addlinespace[2pt]
out-of-order rate$^{\dagger}$ & MuSiQue & $+0.051$ & $+0.058$ & $+0.057$ \\
 & StrategyQA & $-0.011$ & $+0.027$ & $+0.005$ \\
 & 2WikiMultiHopQA & $-0.013$ & $-0.011$ & $-0.012$ \\
\addlinespace[2pt]
\bottomrule
\end{tabular}
\end{table}

\textbf{Development against held out, and the shared cap.} In Table~\ref{tab:app-heldout-coverage} the
trained questioner is above the same model, prompted, at gate-matched spend on all three suites and at
the shared cap of
eight on MuSiQue, within two points of it on the other two. On held-out runs its mean question count
is $4.20$, $3.05$ and $2.32$, the comparator is charged $3.93$, $2.87$ and $2.11$ against its own
$5.90$, $5.96$ and $3.04$, and the comparator's episode was shorter than the charged prefix on $108$,
$69$ and $100$ of $800$ run pairs, where it is charged its own length.

\begin{table}[ht]
\centering
\caption{Required-evidence coverage under the selection gate's rule, the only rule with a development
reading: the trained questioner (two training seeds averaged within the task) at its own stopping
point, the same model, prompted, at its own prefix of that length. Table~\ref{tab:heldout} reads the
same held-out runs at equal spend. Both suites with a
development reading agree with it in sign and in excluding zero, and the MuSiQue gain shrinks from
development to held out, as selection on the development split predicts.
2WikiMultiHopQA has no development reading, so its cell is a first reading rather than a
replication. The splits were scored under different scoring hashes, so across splits this is an
agreement in sign and interval, never a difference of differences. Development intervals are at
$1{,}000$ resamples, held-out ones at $10{,}000$. The last column is the difference at each policy's own
stop under the ceiling of eight.}
\label{tab:app-heldout-coverage}
\resizebox{\linewidth}{!}{\begin{tabular}{llrlrr}
\toprule
suite & split & $\Delta$ & 95\% interval & $n$ tasks & own-stop $\Delta$ \\
\midrule
MuSiQue            & development & $+0.156$ & $[+0.121, +0.204]$ & $132$ & $+0.103$ \\
MuSiQue            & held out    & $+0.128$ & $[+0.102, +0.159]$ & $200$ & $+0.096$ \\
StrategyQA         & development & $+0.080$ & $[+0.059, +0.103]$ & $333$ & $-0.009$ \\
StrategyQA         & held out    & $+0.080$ & $[+0.051, +0.112]$ & $200$ & $-0.012$ \\
2WikiMultiHopQA    & held out    & $+0.079$ & $[+0.052, +0.111]$ & $200$ & $+0.015$ \\
\bottomrule
\end{tabular}}
\end{table}

\textbf{A diversity gate fails under pooling and passes per seed.} The declared diversity statistic,
gated at $0.65$, pools every rollout seed of a task before counting, so a questioner that repeats its
own questions across rollout seeds reads as collapsed, at $0.458$ to $0.600$ held out against $0.827$
to $0.888$ on development, which scores one rollout seed. Grouped by task and rollout seed, as
Table~\ref{tab:app-heldout-gates} prints it, the same runs read $0.805$ to $0.849$: pooling counts
repetition across rollout seeds, not a loss of variety within one.

\begin{table}[ht]
\centering
\caption{Every gated criterion for the trained questioner's two training seeds, development against
held out, under one instrument and estimator. Question diversity is printed per training seed, since
pooling two policies' questions measures something else, and read per task and rollout seed. Question
length fails on every suite and split and makes every overall verdict a fail. Ungated rows are read
at the shared cap of eight, not at equal spend (Appendix~\ref{app:matchedcost}). Coverage at depth two
or more is absent, not zero, on 2WikiMultiHopQA, which has no required need below depth one.
Promptness on MuSiQue falls from $+0.142$ on development, whose interval at $1{,}000$ resamples
excludes zero, to $+0.015$ held out, whose interval spans zero, so held-out promptness there is not
decided and the drop between the splits is not a tested difference.}
\label{tab:app-heldout-gates}
\small
\setlength{\tabcolsep}{4pt}
\resizebox{\linewidth}{!}{\begin{tabular}{llrrrrr}
\toprule
criterion & gated & \multicolumn{2}{c}{MuSiQue} & \multicolumn{2}{c}{StrategyQA} & 2WikiMultiHopQA \\
\cmidrule(lr){3-4}\cmidrule(lr){5-6}\cmidrule(lr){7-7}
 & & dev & held out & dev & held out & held out \\
\midrule
required-evidence coverage & yes & $+0.156$ & $+0.128$ & $+0.080$ & $+0.080$ & $+0.079$ \\
question diversity, seed 1 & yes & $0.827$ & $0.826$ & $0.888$ & $0.849$ & $0.805$ \\
question diversity, seed 2 & yes & $0.841$ & $0.833$ & $0.884$ & $0.830$ & $0.808$ \\
question length            & yes & $17.53$ fail & $17.76$ fail & $22.57$ fail & $22.67$ fail & $12.36$ fail \\
malformed output           & yes & $0.000$ & $0.001$ & $0.003$ & $0.001$ & $0.005$ \\
coverage at depth $\geq 2$ & no  & $+0.097$ & $+0.076$ & $-0.070$ & $-0.005$ & absent \\
breadth                    & no  & $+0.045$ & $+0.084$ & $-0.021$ & $-0.005$ & $+0.021$ \\
promptness                 & no  & $+0.142$ & $+0.015$ & $+0.031$ & $+0.006$ & $+0.086$ \\
\midrule
overall verdict            &     & fail & fail & fail & fail & fail \\
\bottomrule
\end{tabular}}
\end{table}

\textbf{Where the out-of-order excess comes from.} The out-of-order rate counts an edge as violated
when the child's evidence arrives before the parent's, whatever brought it, whether the questioner
asked for the child, skipping the parent, or a question aimed elsewhere, often at the parent,
retrieved the child's paragraph as well. Each violation event on MuSiQue is classified by whether the
question at the turn that surfaced the child's evidence names the child under the matcher's own ask
test. Against the same model, prompted, the trained questioner's asked-for part is $-0.012$ $[-0.042,
+0.014]$ and its incidental part $+0.069$ $[+0.023, +0.115]$. Counting the child as named when all of
its own terms appear gives the same split, $+0.068$ for the incidental part. A looser test, at least
half of the child's terms, moves events from incidental to asked-for and leaves the incidental part at
$+0.040$ $[+0.001, +0.079]$, which holds at fifty thousand resamples under three seeds but crosses
zero at one thousand. The asked-for part spans zero under all three tests. In incidental events the
trained questioner's question named the parent $29$ times, the comparator's $2$. The other two suites show no excess to split.

\textbf{What the disorder costs is not detectable here.} Taking needs out of order should show up as
answers built on unconfirmed premises. Answers are read where each run stops, so this reading takes
complete runs, over the $184$ MuSiQue tasks on which both arms' complete runs define the out-of-order
rate, not the population of the out-of-order row of Table~\ref{tab:app-seeds}, which holds both arms to
the lower question count. Split by whether the trained questioner resolved more out of order than the
comparator on the same task, and paired within task so that difficulty is held fixed, its answer advantage is larger where it resolved more out of order, not
smaller: the strata differ by $+0.081$ $[+0.014, +0.150]$ in answer token overlap, holding at fifty
thousand resamples under three seeds, and by $+0.087$ $[-0.004, +0.181]$ on exact answers, and the
share of tasks with every required need resolved moves the same way, by $+0.140$ $[+0.027, +0.255]$.
That is not evidence that disorder helps, since the strata are chosen by the trained questioner's own
behavior and may simply be the tasks where it reaches more. What it rules out, at this resolution and
on the trained questioner, is the harm the ordering result invites. Each training seed alone leans the
same way on all four measures. The other two suites have $7$ and $6$ more-disordered tasks and are
not read.

\textbf{Independent rollouts, and the population.} The structured-baseline comparison's own rollouts,
drawn independently, agree with Table~\ref{tab:heldout} in sign and in excluding zero: against the same
model, prompted, the trained questioner gains $+0.119$ and $+0.082$ of coverage at equal retrieval spend on MuSiQue and
StrategyQA, where the table prints $+0.112$ and $+0.070$. Every cell reads the $200$ held-out tasks per suite that the eligibility filter admits
(Appendix~\ref{app:metrics}). The train-split contamination
guard returned no violations over all $2{,}400$ of the trained questioner's held-out runs, and $40$
violations over $40$ development-split runs of the trained arm, so the zero is evidence rather than a
guard that cannot fire.
\FloatBarrier
\section{The eight controls}\label{app:controls}

This appendix gives the readings behind the controls of Section~\ref{sec:controls}: what each part of the protocol is
worth to the trained questioner, and which controls can decide anything at all.

\begin{figure}[t]
\centering
\includegraphics[width=0.92\linewidth]{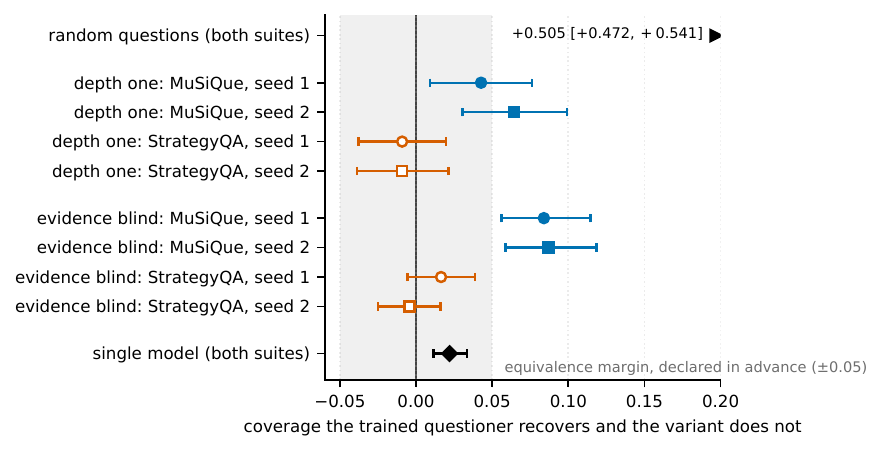}
\caption{What each part of the protocol is worth to the trained questioner, each control read against the trained questioner's
own weights and at the rule its spend requires. Random questions and the single model ask more than the
trained questioner and are read with the control truncated to the trained questioner's spend, pooled over both suites and both
seeds. The depth-one and evidence-blind ablations ask fewer and are read with both arms at the lower
question count, per seed and per suite. A filled point excludes zero, decided at $50{,}000$ resamples
under three seeds for the ablations, and an open point spans zero. Removing what the questions say costs
an order of magnitude more than anything else, and that reading lies beyond the axis, printed at its edge. Sequencing and the evidence each cost MuSiQue on both
seeds and separate nothing on StrategyQA, where the shared budget is one or two questions. The single
model's interval sits inside the grey margin declared in advance in code but excludes zero, which the
declared rule reads as separated. At the shared cap it is equivalent
(Table~\ref{tab:app-controls-pooled}).}
\label{fig:controls}
\end{figure}

\textbf{What a control is here.} Seven comparator arms were registered in advance in code, each with
a written consequence for the case where it matches the questioner it was declared against, the
prompted one, and this appendix reads them against the trained questioner. They are answer length,
compute budget, random questions, depth one, evidence blind, the single model and a determinism check.
An eighth, a state-blind coverage template, was read without a rule, so it can make an unmatched table
read wrong but cannot fire one. Figure~\ref{fig:controls} draws random questions, depth one, evidence
blind and the single model, and Table~\ref{tab:app-controls-pooled} adds answer length, compute
budget and the template. Each declared equivalence margin is $0.05$ of required-evidence coverage except
the answer-length control's $0.03$, tighter because its consequence is an unconditional stop. The
declared verdict reads a control as separated when its interval lies wholly above zero, which is
checked first, as equivalent when the interval lies inside the margin without excluding zero, which
fires the rule, and otherwise as inconclusive. No declared primary endpoint
names the trained questioner, so these contrasts are exploratory as a class and carry intervals and
no probability value. No control arm was run on 2WikiMultiHopQA or on the tool-use domains, so
nothing here extends beyond the two suites read.

\begin{table}[t]
\centering
\caption{The controls against the trained questioner, its two training seeds averaged within each task, pooled over
the held-out splits of MuSiQue and StrategyQA, as required-evidence coverage of the trained questioner minus the
control. The first column charges each control at most the trained questioner's spend, and the second reads both arms at the shared cap
of eight retrieval calls. The three ablations of the questioner's own weights are read for each seed
against its own ablation. Each control takes its own overlap with the trained questioner as its population, which
is why the counts differ. $^{\dagger}$The control is read at the longest prefix of its own run that
stays within the trained questioner's retrieval calls on the same task and rollout seed, and the
trained questioner is read whole. On a pair where the control asks at least as many questions this is
equal spend, both arms at the lower count. On a pair where the control asks fewer, the trained
questioner keeps its extra questions, the gate-matched spend of Appendix~\ref{app:matchedcost}.
$^{*}$The depth-one and evidence-blind ablations ask fewer questions than the
trained questioner, so their first column is not an equal-spend reading, and Figure~\ref{fig:controls} reads them
with both arms at the lower question count.}
\label{tab:app-controls-pooled}
\small
\begin{tabular}{@{}lrrr@{}}
\toprule
control & at most the trained questioner's spend$^{\dagger}$ & shared cap of eight & $n$ \\
\midrule
answer length        & $+0.8848$ $[+0.8660, +0.9020]$ & $+0.8848$ $[+0.8660, +0.9020]$ & 375 \\
compute budget       & $+0.8841$ $[+0.8654, +0.9013]$ & $+0.8841$ $[+0.8654, +0.9013]$ & 375 \\
random questions     & $+0.5049$ $[+0.4723, +0.5408]$ & $+0.3648$ $[+0.3333, +0.4021]$ & 376 \\
state-blind template & $+0.2439$ $[+0.2177, +0.2718]$ & $+0.1892$ $[+0.1614, +0.2171]$ & 400 \\
depth one$^{*}$      & $+0.1448$ $[+0.1242, +0.1668]$ & $+0.1380$ $[+0.1176, +0.1596]$ & 400 \\
evidence blind$^{*}$ & $+0.0701$ $[+0.0529, +0.0861]$ & $+0.0369$ $[+0.0200, +0.0525]$ & 400 \\
single model         & $+0.0220$ $[+0.0114, +0.0335]$ & $-0.0028$ $[-0.0129, +0.0075]$ & 400 \\
\bottomrule
\end{tabular}
\end{table}

\noindent Per suite, the trained questioner leads random questions at equal spend by $+0.3474$ $[+0.3033, +0.3910]$ on
MuSiQue, over the $176$ tasks the control keeps, and by $+0.6325$ $[+0.5923, +0.6757]$ on StrategyQA,
over $200$.

With both arms at the lower question count, on MuSiQue the trained questioner leads its depth-one
ablation by $+0.043$ and $+0.064$ for its two seeds and its evidence-blind ablation by $+0.084$ and
$+0.087$, every interval excluding zero, and on StrategyQA neither separates.

\textbf{The single model: equivalent at the shared cap, separated at equal spend.} The single-model
variant, one model playing questioner and drafter, reads $-0.0028$ $[-0.0129, +0.0075]$ against the
trained questioner at the shared cap of eight, inside the margin of $0.05$ without excluding zero, so the declared
verdict there is equivalent and its consequence applies: the contribution is the Q\&D algorithm rather
than the two-model split. At equal spend it trails the split by $+0.0220$ $[+0.0114, +0.0335]$, and per
suite by $+0.0257$ $[+0.0145, +0.0398]$ on MuSiQue and $+0.0186$ $[+0.0029, +0.0380]$ on StrategyQA.
Each of those intervals lies inside the margin but excludes zero, which the declared verdict, checking
separation first, reads as separated: at equal spend the split buys a small, detectable amount of
coverage.

\textbf{Three controls that cannot decide.} The answer-length and compute-budget arms ask nothing, so
their coverage is zero on every task and their deltas of $+0.885$ and $+0.884$ are true and are not
evidence, which leaves those two explanations unfalsified rather than ruled out, and the determinism
check reissues a recorded run's own questions, so its delta against that run is zero by construction
and it has no row here. The question-length control of Section~\ref{sec:length} is a separate
instrument.
\FloatBarrier
\section{Stopping in full}\label{app:stopping}

This section supports the stopping result of Section~\ref{sec:stopping} (Table~\ref{tab:mechanism}) and the
call-budget sweep of Figure~\ref{fig:frontier}a: why filtered imitation teaches no persistence where it is needed, how the
preference stage restores it and at what cost, the answer-node arm, and where the evidence gain is lost
before the answer.

\begin{figure}[t]
\centering
\includegraphics[width=0.6\linewidth]{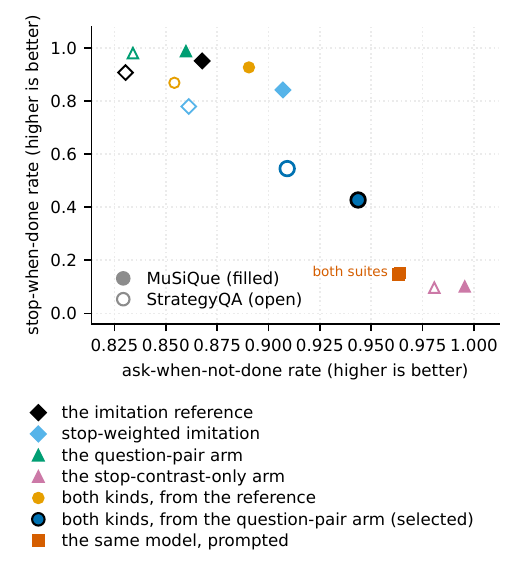}
\caption{The trade every training arm makes between persistence and stopping: each marker is one
trained arm at its rate of asking when the task is not done and of stopping when it is, with the same
model, prompted, for reference. Arms trained on one kind of contrast sit at opposite corners, and the
trained questioner (larger marker, two training seeds pooled) sits between them, asking at an
unfinished state more often and stopping at a finished state less often than every other trained arm
but the one trained on stop contrasts alone, with a coverage gain within $0.012$ of the largest on
MuSiQue and within $0.005$ on StrategyQA. Shape marks the kind of training and color the arm. A filled
marker is MuSiQue and an open one StrategyQA, and the same model's two suites nearly coincide.
Development split, rates over the decision points each policy reaches.}
\label{fig:stopping}
\end{figure}

\textbf{Two criteria fixed in advance that the trained questioner does not clear.} The selection rule
asked for a rise of at least $0.03$ in the rate of asking at an unfinished state, a rate of stopping at
a finished state of at least $0.85$, and a non-inferiority test on question length. The trained
questioner clears persistence on both suites, its rate of asking at an unfinished state up by $+0.076$
and $+0.079$ against the reference, and fails the other two on both: it stops at a finished state at
$0.427$ and $0.545$, and its questions run four and a half and nine and a half words longer than the
comparator's, each relative interval wholly above the margin (Table~\ref{tab:app-stop-length}).

\begin{table}[t]
\centering
\caption{The gated one-sided length non-inferiority test on StrategyQA, with the MuSiQue verdict
beside it: mean question words against the same model, prompted, where the upper bound of the relative
interval may not exceed a margin of $0.10$. All four trained arms fail on StrategyQA and the one from the question-pair arm fails
MuSiQue as well. The reference passes by $0.007$ of margin, so the criterion decides more than its
precision can bear.}
\label{tab:app-stop-length}
\small
\resizebox{\linewidth}{!}{\begin{tabular}{@{}lrrcccc@{}}
\toprule
arm & words & comparator & relative 95\% CI & StrategyQA & MuSiQue \\
\midrule
imitation reference       & 13.80 & 12.99 & $[+0.031,+0.093]$ & pass & pass \\
stop contrasts alone      & 16.59 & 12.99 & $[+0.256,+0.297]$ & fail & pass \\
both kinds                & 14.31 & 12.99 & $[+0.072,+0.129]$ & fail & pass \\
both kinds, from the question-pair arm & 22.57 & 12.99 & $[+0.705,+0.768]$ & fail & fail \\
stop rows down-weighted to $0.1$ & 14.36 & 12.99 & $[+0.079,+0.134]$ & fail & pass \\
\bottomrule
\end{tabular}}
\end{table}

\textbf{Imitation leaves the hard states without a lesson, and a contrast supplies one.} The imitation
exporter writes a stop where the required evidence was in hand, the best sampled question where
evidence was missing and that question clears the acceptance floor, and nothing where none clears it.
The last branch suits an imitation target, since a state every sample missed says more about the
sampler than the state, but it drops precisely the states whose lesson is that stopping is wrong, so the
policy trained on the rest has no feature that separates a hard state from a finished one. It refuses
$5{,}030$ states beside $14{,}302$ exported rows on the export the reference trained on, and $24{,}538$
beside $43{,}837$ on the pooled export, which keeps only runs of the same model, prompted, with no
cohort filter, and is not the widest export of Appendix~\ref{app:training}, which admits every arm. A preference need not name a good question, only that asking beat
stopping, which gold coverage settles on its own, so at every unfinished state, the refused ones
included, the export places the best available question against stopping, ordered without consulting
the question's text: $18{,}919$ of the file's $36{,}349$ pairs. Three compositions were gated against one
comparator (Figure~\ref{fig:stopping}): those contrasts alone, and both kinds together, $31{,}473$
pairs, fitted to the imitation reference or to the policy already trained on question pairs.

\begin{table}[t]
\centering
\caption{Training composition against behavior on the development split. Persistence is the rate of
asking when the task is not done and stop the rate of stopping when it is, per decision point the
policy itself reaches. Coverage is against the same model, prompted, at gate-matched spend, with
a task-clustered interval. Diversity is the within-task distinct-trigram share, against a gate floor of
$0.65$ and untrained values of $0.380$ and $0.458$. Asks is questions per episode. Every arm but
the last of each block is one training run, and the last averages two, scored in a separate pass
against the same comparator runs.}
\label{tab:app-stop-composition}
\footnotesize
\resizebox{\linewidth}{!}{\begin{tabular}{@{}lrrrrrrr@{}}
\toprule
composition & pairs & persist. & stop & $\Delta$ cov. & 95\% CI & div. & asks \\
\midrule
\multicolumn{8}{l}{\textbf{MuSiQue}, 132 tasks} \\
imitation reference    & none     & 0.868 & 0.951 & $+0.049$ & $[+0.002,+0.097]$ & 0.830 & 3.05 \\
stop contrasts alone   & 18{,}919 & 0.996 & 0.101 & $-0.005$ & $[-0.056,+0.046]$ & 0.447 & 6.95 \\
both kinds             & 31{,}473 & 0.890 & 0.926 & $+0.068$ & $[+0.021,+0.117]$ & 0.801 & 3.36 \\
both kinds, from the question-pair arm & 31{,}473 & 0.944 & 0.427 & $+0.156$ & $[+0.121,+0.204]$ & 0.834 & 4.24 \\
\midrule
\multicolumn{8}{l}{\textbf{StrategyQA}, 333 tasks} \\
imitation reference    & none     & 0.830 & 0.907 & $+0.080$ & $[+0.052,+0.111]$ & 0.845 & 1.34 \\
stop contrasts alone   & 18{,}919 & 0.981 & 0.097 & $-0.017$ & $[-0.039,+0.004]$ & 0.445 & 5.75 \\
both kinds             & 31{,}473 & 0.854 & 0.869 & $+0.084$ & $[+0.056,+0.111]$ & 0.845 & 1.47 \\
both kinds, from the question-pair arm & 31{,}473 & 0.909 & 0.545 & $+0.080$ & $[+0.059,+0.103]$ & 0.886 & 2.13 \\
\bottomrule
\end{tabular}}
\end{table}

\textbf{What the composition decides.} The stop contrasts move stopping only when they dominate the
objective or meet a policy that already asks well (Table~\ref{tab:app-stop-composition}). Alone they raise persistence and buy nothing else:
the coverage point estimate is negative on both suites with neither interval excluding zero,
within-task diversity is the lowest of any trained arm measured here, and more than twice the reference's questions recover no more required evidence. Fitted to
the imitation reference beside question pairs, where they are still the majority, they leave stopping
when done intact and return almost none of the persistence gain. Fitted to the
policy already trained on question pairs from the reference, the same file moves both cells and
gains more than twice as much MuSiQue coverage as when fitted to imitation, so the final stage moves the
policy toward persistence. That starting policy has no development reading here. Held out, the final
stage and the policy it starts from gain $+0.1284$ and $+0.1177$ on MuSiQue
(Table~\ref{tab:app-heldout-methods}), a point ordering and not a tested difference. The question-pair
row of Table~\ref{tab:mechanism} is a different checkpoint, fitted from the pooled imitation export: it
gains $+0.167$ $[+0.127, +0.215]$ and $+0.084$ $[+0.057, +0.113]$ and leaves both stopping rates about
where imitation left them. Down-weighting the stop rows in the imitation loss cannot put back the refused
states and is not free: with the stop rows down-weighted to a tenth, stopping at a finished state falls
to $0.7792$ on StrategyQA, below the selection floor.

\textbf{Against GPT-OSS-120B, the trained questioner keeps the stopping half of the calibration and
not the adverse half.} On the same instrument and held-out split, it stops at $0.64$ to
$0.92$ of its own finished states against that model's $0.23$ to $0.81$, higher by $+0.251$, $+0.430$
and $+0.116$, every interval excluding zero. It asks when evidence is still missing more often than that
model on MuSiQue, by $+0.035$ $[+0.017, +0.059]$, less often on StrategyQA, by $-0.067$
$[-0.088, -0.050]$, and indistinguishably on 2WikiMultiHopQA, at $-0.007$ $[-0.022, +0.005]$, so it is
lower only on the suite where it is also lower than the same model, prompted. Each rate is read only
where a policy reached a decision point of the matching kind, a population selected on both policies'
trajectories, and rates rather than counts are compared because the two take very different numbers of
turns.

\textbf{The call-budget sweep of Figure~\ref{fig:frontier}a.} On MuSiQue the trained questioner leads
GPT-OSS-120B by $+0.102$ $[+0.074, +0.134]$ at the cheapest ceiling, where both spend about $3.2$ calls,
and by $+0.064$ $[+0.037, +0.095]$ at the most generous, spending $5.9$ calls against $12.1$. Its
coverage at a ceiling of four is above that model's at twenty-four, by $+0.041$ $[+0.013, +0.071]$. It
leads the same model, prompted, by $+0.093$ to $+0.096$ at every ceiling, and each training seed alone
is ahead of both comparators at every ceiling.

\begin{figure}[t]
\centering
\includegraphics[width=\linewidth]{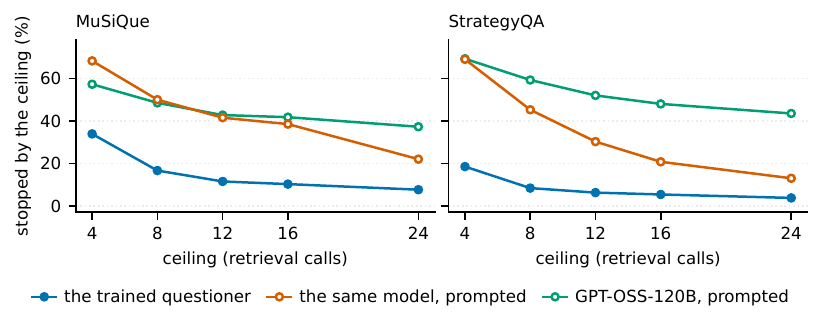}
\caption{Who decides when a run ends: the share of units (one task at one rollout seed) that the
retrieval ceiling rather than the policy stopped, at five ceilings across a sixfold budget range. From
the smallest ceiling to the largest the trained questioner is stopped on $34\%$ to $8\%$ of MuSiQue
units and $19\%$ to $4\%$ of StrategyQA units, below the same model, prompted, and GPT-OSS-120B,
prompted, at every ceiling on both suites, which are stopped on $13\%$ to $69\%$. The same
policy also stops on unfinished work (Section~\ref{sec:stopping}).}
\label{fig:ceiling}
\end{figure}

On StrategyQA no difference from the same model, prompted, is decided at any ceiling, $+0.008$ to
$+0.012$ with every interval spanning zero, on $2.5$ to $3.9$ calls against $3.6$ to $10.0$. No
difference from GPT-OSS-120B is decided at the cheapest ceiling, $+0.009$ $[-0.016, +0.037]$, and the
trained questioner trails that model from the second ceiling upward, by $-0.027$ $[-0.050, -0.002]$ at
a ceiling of eight and $-0.056$ $[-0.079, -0.035]$ at twenty-four, where that model spends two to three
and a half times as many calls. From the smallest ceiling to the largest the trained questioner gains
$+0.023$ on MuSiQue and $+0.015$ on StrategyQA, GPT-OSS-120B $+0.062$ and $+0.080$. On FRAMES, which
has no need graph and is read on answers, no difference from the same model, prompted, is decided at
any ceiling, the point estimates running $+0.010$ to $+0.019$, and none from GPT-OSS-120B up to a
ceiling of twelve. The trained questioner trails GPT-OSS-120B at sixteen and twenty-four, by $-0.030$
and $-0.033$, where that model spends two to two and a half times as many calls. Figure~\ref{fig:ceiling} shows how often the cap rather than the policy
ends a run.

\textbf{The answer-node arm against its own control.} The repair that follows from a policy that
stops with the answer-bearing need uncovered, as the trained questioner still does on StrategyQA, is
to supervise stopping on that node itself. An imitation arm trained that way covers that node less often than
its control, refreshed by the same imitation recipe without the conditioning, by $-0.095$ $[-0.148, -0.048]$, $-0.075$ $[-0.125, -0.033]$
and $-0.293$ $[-0.358, -0.233]$ on the three suites ($200$ paired tasks each), every interval excluding
zero and every sign stable across resample counts and seeds. Against the same model, prompted, the same
endpoint reads $-0.128$, $-0.175$ and $-0.370$.

\textbf{The node's depth tests the alternative reading.} That reading is that the deficit measures needs
not yet nameable rather than stopping that was wrong. On 2WikiMultiHopQA it inverts: the deficit is
larger where the answer node was nameable from the task, $-0.394$ $[-0.500, -0.303]$, than behind a
prerequisite, $-0.203$ $[-0.293, -0.132]$. On StrategyQA the nameable stratum is too small to decide,
at $-0.043$ $[-0.114, +0.014]$ over $70$ tasks, and MuSiQue has none by construction, no graph in it
having a required terminal need at depth zero. So the stopping defect is not a mislabeled target: handed
the answer node directly, an imitation policy does worse on it, which locates the problem in what it
generalizes at inference rather than in what it was taught to predict.

\subsection{Where the evidence gain is lost before the answer}\label{app:answerloss}

At each arm's own stop under the ceiling of eight, with the trained questioner's two training seeds
averaged within the task, only MuSiQue has extra answer-node coverage to convert
(Table~\ref{tab:app-answerloss}), and the gain it implies, from the comparator's own rate of answering
well with and without that node, is below the detectable size. The declared test of a stronger reader, Claude Opus 5 given the
retrieved paragraphs uncut and told to answer only from them, is decided in none of twelve cells, at each arm's
own stop and at equal spend alike. On StrategyQA that reader answers that the evidence does not settle
the question on about $45\%$ of states, becoming a strict entailment reader rather than a measure
of headroom. The drafter's cut at $1{,}200$ characters hides no answer on MuSiQue ($0$ of $212$
answer-bearing paragraphs) and $6$ of $174$ on 2WikiMultiHopQA, and is not read on StrategyQA, whose
answers are yes or no. The frozen answerer is not deterministic either: redrawing it on byte-identical
requests turns the MuSiQue answer-overlap difference into $+0.036$ $[+0.002, +0.069]$.

\begin{table}[h]
\caption{The answer gain the trained questioner's evidence would imply, the gain observed, and a stronger reader. The node column is the difference in covering the answer-bearing need. Answer token overlap, the trained questioner minus the same model, prompted, $200$ held-out tasks per suite, $10{,}000$-resample intervals. The minimum detectable effect is at $80\%$ power and two-sided $0.05$ from the observed paired spread. The last column is the declared primary, the difference between that difference read under a Claude Opus 5 evidence-only reader and read under the frozen answerer, at each arm's own stop. Intervals here cluster on the MuSiQue question template, and the point moves with them: the estimate is the mean over clusters, so here each template weighs the same whatever its number of tasks, where Table~\ref{tab:heldout}, clustered on the task, weighs each task the same. That is why the observed MuSiQue cell reads $+0.026$ here and $+0.024$ there for the same runs.}
\label{tab:app-answerloss}
\begin{center}\footnotesize\setlength{\tabcolsep}{5pt}
\begin{tabular}{@{}lccccc@{}}
\toprule
suite & node & implied & observed & detectable & stronger reader \\
\midrule
MuSiQue & $+0.097$ & $+0.043$ & $+0.026$ {\scriptsize $[-0.006, +0.060]$} & $0.047$ & $+0.026$ {\scriptsize $[-0.015, +0.067]$} \\
StrategyQA & $-0.005$ & $-0.001$ & $-0.003$ & $0.056$ & $-0.001$ {\scriptsize $[-0.053, +0.049]$} \\
2WikiMultiHopQA & $+0.001$ & $+0.001$ & $+0.031$ & $0.053$ & $+0.007$ {\scriptsize $[-0.018, +0.035]$} \\
\bottomrule
\end{tabular}
\end{center}
\end{table}
\FloatBarrier
\section{Transfer in full}\label{app:transfer}

This appendix supports the transfer result of Section~\ref{sec:transfer} (Figure~\ref{fig:tau2}): the trained questioner's gain in task success on a
tool-use benchmark, the follow-up turns that were its declared primary endpoint, and its comparison with
GPT-OSS-120B. It also reports two campaigns of the questioner-drafter protocol, the questioner-drafter
split of Q\&D, run with prompted models and no training: with Claude Sonnet 5 in every role it needs
fewer follow-up turns on retail, a difference not decided on airline, and with GPT-OSS-120B in every
role it decides nothing on banking. Neither campaign has a trained arm, so neither bears on the
trained questioner's declared primary endpoint.

\textbf{The trained questioner, at equal and enforced budgets.} We compared the trained questioner
from two training seeds with the same model, prompted, on a tool-use benchmark with a simulated
customer \citep{barres2025tau2}, on its retail and airline domains. A fork point is a prefix of a
recorded dialogue from which a run continues, and each domain has thirty-four, drawn from $25$ retail
and $17$ airline tasks. Each fork point was run by three policies, the same model, prompted, and the
two training seeds, at three rollout seeds and under two prompt variants, $1{,}224$ units over the two
domains, every one completed. Both prompt variants add to the questioner's question-answering prompt one
paragraph saying that a question reaches the store's records and never the customer. The base prompt keeps
the question-answering stopping rule, to stop once every required need is resolved, and the stop prompt
replaces it with one written for tool use, to stop once the evidence names the records the task's next
action needs, since the drafter acts only after the questioner stops. The two are read separately and
never pooled. Every difference is read over tasks, with the pairs inside a task
averaged first. In both arms the harness enforced sixteen
questions and sixteen tool calls separately during the dialogue, the questioner ran with its reasoning
mode off, and the drafter saw each evidence record whole. The analysis rules were fixed in writing before the data each rule governs.

Retail success rose in every seed and prompt variant (Table~\ref{tab:tau2-trained}). Airline success
rose in each pooled cell and in both variants for the second seed, and the first seed's airline
cells are undecided, with two or three of seventeen tasks better per seed and at most one worse.
Follow-up turns, the declared primary endpoint,
fell in every pooled estimate, and no pooled interval excludes zero, so the endpoint is not
established.

On the two confirmatory question measures, the gold-object hit rate, the share of questions whose
retrieved records include one the task's gold actions touch, rose by $0.185$ to $0.265$ (in retail on
twenty-one of twenty-five tasks, with none against), and gold-object recall rose by $0.324$ to
$0.512$, while the number of questions fell slightly.

Two further measures, declared after one interim look and so exploratory, point the same way in every
domain and prompt (Table~\ref{tab:tau2-ask-quality}): more questions retrieve at least one record, and
fewer exactly repeat an earlier question. A looser repeat measure falls in retail. In airline no
difference in it is decided, and about half of the trained questioner's questions nearly repeat an
earlier one, $0.497$ against $0.421$ under the base prompt.

In a worked example chosen by a rule fixed before any transcript was opened (retail task 15), no
question of the same model, prompted, retrieved any of the five records the task needs, and it failed.
Under the base prompt ten questions of the first trained seed each retrieved at least one of those records, and
that seed succeeded under both prompt variants.

\begin{table}[h]
\centering
\small
\caption{The trained questioner against the same model, prompted, on the tool-use benchmark, at equal, separately enforced budgets. Differences are trained minus prompted, over tasks, with bias-corrected intervals at $10{,}000$ resamples.}
\label{tab:tau2-trained}
\footnotesize
\begin{tabular}{@{}llcccccc@{}}
\toprule
 & & \multicolumn{3}{c}{Task success} & Follow-up turns & Hit rate & Recall \\
domain & prompt & Seed 1 & Seed 2 & Pooled & Pooled & Pooled & Pooled \\
\midrule
retail & base & $+0.213$ & $+0.193$ & $+0.203$ & $-0.660$ & $+0.254$ & $+0.512$ \\
 & & & & {\scriptsize $[+0.100, +0.357]$} & {\scriptsize $[-1.853, +0.387]$} & & \\[2pt]
retail & stop & $+0.233$ & $+0.167$ & $+0.200$ & $-0.737$ & $+0.265$ & $+0.504$ \\
 & & & & {\scriptsize $[+0.100, +0.340]$} & {\scriptsize $[-2.343, +0.870]$} & & \\[2pt]
airline & base & $+0.059$ & $+0.078$ & $+0.069$ & $-0.059$ & $+0.185$ & $+0.368$ \\
 & & & & {\scriptsize $[+0.010, +0.196]$} & {\scriptsize $[-1.064, +0.941]$} & & \\[2pt]
airline & stop & $+0.029$ & $+0.098$ & $+0.064$ & $-0.441$ & $+0.224$ & $+0.324$ \\
 & & & & {\scriptsize $[+0.005, +0.201]$} & {\scriptsize $[-1.643, +0.985]$} & & \\[2pt]
\bottomrule
\end{tabular}
\end{table}

\begin{table}[h]
\centering
\small
\caption{Exploratory question measures on the tool-use benchmark, declared after one interim look and
so exploratory whatever they show. Differences are trained minus prompted, over tasks ($25$ retail,
$17$ airline), with the two training seeds pooled within each task and bias-corrected intervals at
$10{,}000$ resamples, not read at $50{,}000$. A near-exact repeat shares at least four content words
with an earlier question in the same dialogue.}
\label{tab:tau2-ask-quality}
\footnotesize
\begin{tabular}{@{}llccc@{}}
\toprule
domain & prompt & retrieved a record & exact repeat & near-exact repeat \\
\midrule
retail & base & $+0.582$ & $-0.613$ & $-0.205$ \\
 & & {\scriptsize $[+0.447, +0.693]$} & {\scriptsize $[-0.675, -0.527]$} & {\scriptsize $[-0.288, -0.126]$} \\[2pt]
retail & stop & $+0.587$ & $-0.609$ & $-0.237$ \\
 & & {\scriptsize $[+0.450, +0.701]$} & {\scriptsize $[-0.671, -0.522]$} & {\scriptsize $[-0.313, -0.163]$} \\[2pt]
airline & base & $+0.382$ & $-0.516$ & $+0.076$ \\
 & & {\scriptsize $[+0.258, +0.512]$} & {\scriptsize $[-0.605, -0.427]$} & {\scriptsize $[-0.065, +0.249]$} \\[2pt]
airline & stop & $+0.394$ & $-0.487$ & $+0.027$ \\
 & & {\scriptsize $[+0.256, +0.538]$} & {\scriptsize $[-0.576, -0.393]$} & {\scriptsize $[-0.100, +0.202]$} \\[2pt]
\bottomrule
\end{tabular}
\end{table}

\textbf{Against GPT-OSS-120B, prompted.} We ran the same comparison with GPT-OSS-120B, prompted, as
the comparator's questioner, under the same harness, budgets and roles, declared before any of its
units ran. As questioner, GPT-OSS-120B came within a tenth of its token limit on $0.6\%$ of calls. In
retail the trained questioner cut follow-up turns and raised success in both prompt variants
(Table~\ref{tab:tau2-vs-120b}). The reduction in follow-up turns holds for the pooled seeds and for
the first training seed, and the second seed points the same way without excluding zero. Success was
higher on ten of twenty-five tasks and lower on one, in each variant. Per seed, the first seed's base-prompt follow-up
reduction passes a Holm-corrected sign test at $0.05$ and no success cell does, while the
intervals decide several. The trained questioner's questions retrieved a record the gold actions
touch more often, by $0.099$ and $0.130$ per question. In an exploratory analysis, a larger share of
its questions retrieved at least one record than GPT-OSS-120B's in all four settings, $0.595$
against $0.323$ and $0.602$ against $0.314$ in retail and $0.555$ against $0.433$ and $0.548$ against
$0.409$ in airline, while in airline under the stop prompt it also repeated earlier questions more
often. In airline neither follow-up turns nor success showed a decided difference, and success
differed on three of seventeen tasks, one higher and two lower under the base prompt and the reverse
under the stop prompt. There the trained questioner asked between one
and one and a half more questions per continuation than GPT-OSS-120B. Two units in which GPT-OSS-120B
asked nothing were read as zero spend under a rule
fixed before any comparison was read.

\begin{table}[t]
\centering
\small
\caption{The trained questioner against GPT-OSS-120B, prompted, on the tool-use benchmark at equal, separately enforced budgets. Differences are trained minus prompted, over tasks, with the two training seeds pooled within each task, and 95\% intervals. A negative difference in follow-up turns means fewer turns from the simulated user. Bold: decided, every interval from $50{,}000$ resamples at three bootstrap seeds excludes zero. Hit rate and questions are secondary point differences.}
\label{tab:tau2-vs-120b}
\begin{tabular}{@{}llcccc@{}}
\toprule
domain & prompt & Follow-up turns & Task success & Hit rate & Questions \\
\midrule
retail & base & $\mathbf{-1.60}$ {\scriptsize $[-3.05, -0.19]$} & $\mathbf{+0.170}$ {\scriptsize $[+0.070, +0.310]$} & $+0.099$ & $-0.87$ \\
retail & stop & $\mathbf{-1.75}$ {\scriptsize $[-3.18, -0.33]$} & $\mathbf{+0.127}$ {\scriptsize $[+0.043, +0.240]$} & $+0.130$ & $+0.19$ \\
airline & base & $+0.75$ {\scriptsize $[-0.39, +1.75]$} & $0.000$ {\scriptsize $[-0.039, +0.049]$} & $+0.052$ & $+1.09$ \\
airline & stop & $-0.07$ {\scriptsize $[-1.42, +1.21]$} & $+0.034$ {\scriptsize $[-0.005, +0.147]$} & $+0.067$ & $+1.40$ \\
\bottomrule
\end{tabular}
\end{table}

\begin{figure}[t]
\centering
\includegraphics[width=\linewidth]{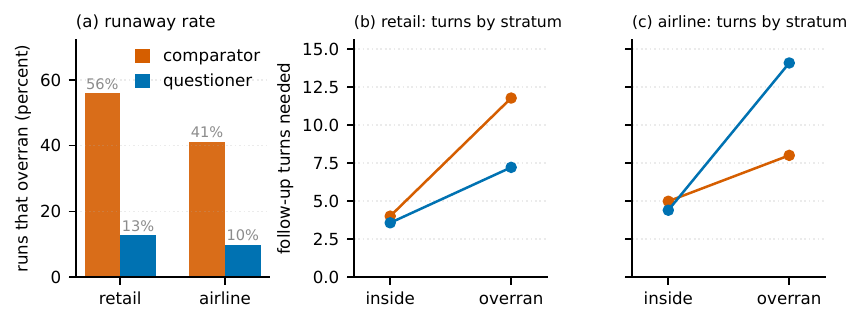}
\caption{What transfers on a user-facing benchmark, and what does not. (a) The share of runs that
overran the recorded budget: the questioner-drafter protocol's dialogues run away roughly four times less often, on
both domains. (b) On retail the protocol needs fewer follow-up turns in both populations, almost all of
the effect in the overrun one: the inside-budget levels sit $0.44$ turns apart. (c) On airline the rate
still favors the protocol, $10$ runs against $42$, but its few runaways are longer, so the mean cannot
separate and only retail's interval excludes zero. The arms are the questioner-drafter protocol and a self-ask
decomposition comparator, both run with Claude Sonnet 5 in every role, on an exploratory-flagged
population that nothing here was trained on. The inside-budget levels describe two populations and
are not a second estimate of the effect.}
\label{fig:transfertail}
\end{figure}

\textbf{The questioner-drafter protocol with Claude Sonnet 5.} This is a separate fork campaign with no
trained arm, so its result concerns the protocol and not the trained questioner's declared endpoint above. With Claude Sonnet 5 in every role on both sides, the
questioner-drafter protocol was compared with a self-ask decomposition comparator on forked dialogue prefixes,
$34$ fork points per domain at three seeds, $204$ runs per domain, paired at every fork point, with the
follow-up turns the simulated customer supplies as the endpoint. The runs are flagged exploratory, so
the contrast is read as an interval with no probability value. Both arms saw only the first $1{,}200$
characters of each tool result, so the comparison is symmetric but read on truncated records. It rests
on the recorded runs of that fork campaign, which drew five evidence units per search. Every spend total for this environment is a
lower bound. On retail the protocol needs $4.47$ fewer follow-up turns $[-6.28, -3.04]$, clustered on
the $32$ recorded dialogues, and every other clustering unit also excludes zero. The difference is in
how often a dialogue runs away rather than in the typical dialogue (Figure~\ref{fig:transfertail}): the
comparator overran a retrieval cap of $16$, charged only after the dialogue, on $57$ of its $102$ retail runs against $13$ of the protocol's. In both arms every retail
question addresses the retrieval index and none the customer.

On airline the interval contains zero at every clustering unit (Figure~\ref{fig:transfertail}c). In
airline held-out runs the only arm permitted to address the customer, a prompted questioner, almost
never does so, an exploratory reading: $8$ of its $1{,}243$ questions go to the customer. Banking, a
third domain, is the banking customer-service domain with a document knowledge base that the
benchmark's public code repository distributes. It was run as a separate live campaign of three arms
with GPT-OSS-120B, prompted, in every role and no trained arm: the questioner-drafter protocol, its
single-model control and a no-question baseline. All $1{,}472$ recorded ask turns
address the retrieval index and none the customer. Over the $83$ of $97$ tasks that survive in all
three arms, the questioner-drafter protocol's difference from the no-question baseline is
$-1.29$ $[-3.51, +1.13]$ follow-up turns over tasks and $-0.78$ $[-2.99, +2.04]$ over its $18$
templates, the benchmark's declared unit. Every interval on every contrast covers zero, and every
detectable floor exceeds the observed difference by between $2.03$ and $13.78$ times, so this null is
a statement about the population's size and not a demonstration that the protocol does nothing.

\textbf{What a simulated customer leaves untested.} Users who are heterogeneous, impatient or working
from shifting instructions \citep{yu2026wildtool,yang2026rutbench}, long-running personalized
interaction \citep{chen2026vitabench2} and interactive coding sessions \citep{wu2026swetogether} have
their own benchmarks, and whether a need graph survives a user who does not behave is open.

\FloatBarrier
\FloatBarrier
\section{Label validity}\label{app:validity}

This appendix supports the check of the pair-ordering rule against a model panel and two people
(Section~\ref{sec:controls}) and the statement that prerequisite edges rarely gate retrieval
(Section~\ref{sec:limits}).

\subsection{The model panel, and one replacement}

\textbf{Pairs are ordered by a rule, and the rule is checked against readers.} Three model families
rated blinded pairs ordered by the rule, with planted items whose answer is fixed in advance. One
member, the model that generated the corpus it rated, agreed with the rule barely more often than chance
and was replaced by a model from a fourth family presented in mirrored option order. The other two
members' votes were not re-collected. The replacement reached the pairwise orderings only: $4{,}476$ of
the $4{,}610$ verdicts of the instrument that ranks every candidate at a state carry the replaced
member.

\begin{table}[ht]
\centering
\caption{chance-corrected agreement on the axis that decides a pair, slot coding, sided votes}
\label{tab:app-kappa-headline}
\small
\begin{tabular}{lrlr}
\toprule
comparison & raw & Cohen's $\kappa$ [95\% clustered] & $n$ / clusters \\
\midrule
the two retained members, both panels   & 77.8\% & 0.555 [0.481, 0.624] & 2{,}508 / 201 \\
first retained with replaced            & 65.2\% & 0.304 [0.244, 0.364] & 3{,}156 / 211 \\
second retained with replaced           & 58.2\% & 0.163 [0.053, 0.259] & 2{,}648 / 202 \\
first retained with replacement         & 79.0\% & 0.581 [0.522, 0.638] & 2{,}596 / 202 \\
second retained with replacement        & 84.1\% & 0.682 [0.629, 0.734] & 2{,}223 / 197 \\
\midrule
rule with first retained                & 61.5\% & 0.230 [0.167, 0.294] & 3{,}480 / 215 \\
rule with second retained               & 67.5\% & 0.350 [0.265, 0.437] & 2{,}931 / 207 \\
rule with replaced                      & 52.2\% & 0.044 [$-$0.027, 0.110] & 4{,}312 / 216 \\
rule with replacement                   & 65.5\% & 0.310 [0.236, 0.382] & 3{,}250 / 208 \\
rule with panel majority, original      & 64.0\% & 0.279 [0.217, 0.341] & 3{,}218 / 213 \\
rule with panel majority, revised       & 68.1\% & 0.361 [0.289, 0.430] & 2{,}986 / 207 \\
\bottomrule
\end{tabular}
\normalsize
\end{table}

\textbf{Agreement with the rule, corrected for chance and clustered at the question.} Every
coefficient \citep{cohen1960kappa,krippendorff2004content} is computed on the presentation slot, a
per-item coin flip every party is blind to, since coding a pair by which side the rule chose makes the
coefficient identically zero. Intervals are clustered over $219$ questions and run $2.2$ to $2.7$ times
wider than row-level ones. Against the rule (Table~\ref{tab:app-kappa-headline}), the replaced member's
interval contains zero and the retained members sit at $0.230$ and $0.350$, so the panel is two
fair-to-moderate raters and one rater not distinguishable from chance. The rule's agreement with the panel majority, $64.0\%$ on the $3{,}218$ pairs the original
panel ordered and $68.1\%$ on the $2{,}986$ the revised panel ordered, is not over a fixed
population. The pooled $0.279$ averages a stratum at $0.386$
with one at or below chance, the pairs where the chosen side resolved no required need, at $-0.095$
$[-0.241, 0.044]$ and $-0.028$ $[-0.205, 0.142]$ under the two panels.

\subsection{Two human raters}

\textbf{Both people passed every hidden check, and they sit where the panel sits.} A blinded bundle of
$113$ items, a hundred real pairs from MuSiQue with ten foils and three attention checks, followed the
key design the model raters saw. Each of the two raters answered every item. Both raters scored $13$ of $13$ on the planted items and chose the first slot on $52.4\%$
$[40.3, 64.2]$ and $50.0\%$ $[38.4, 61.6]$ of the items they decided, so neither answered from position,
which matters because the slot order is shared across raters. They agree with the rule on $65.1\%$
$[52.8, 75.7]$ of $63$ decided items and $67.6\%$ $[55.8, 77.6]$ of $68$, against the panel
majority's $68.1\%$ on the $72$ of these items it decided, both intervals containing the panel's
value, so the rule's agreement with its
raters is not a consequence of the raters being models.

\begin{table}[h]
\centering
\caption{Agreement between the two people on the human anchor's items, per question asked of them,
with chance-corrected agreement.}
\label{tab:app-human}
\begin{tabular}{llll}
\hline
question asked of the rater & items & raters agree & chance-corrected \\
\hline
does candidate A reach an unstated need & $98$ & $87.8\%$ & $+0.756$ \\
does candidate B reach an unstated need & $92$ & $88.0\%$ & $+0.761$ \\
which candidate is the better move & $56$ & $67.9\%$ & $+0.241$ \\
\hline
\end{tabular}
\end{table}

\noindent\textbf{The two labels split sharply.} The reaching judgment, whether a candidate reaches a
need the task never stated, has a human anchor at $0.76$, and the preference ordering does not. The two retained model raters show
the same split on $280$ matched pairs in one forced-pair format, $0.878$ $[0.798, 0.939]$ on which
candidate reaches further past what the user stated against $0.556$ $[0.373, 0.709]$ on which is the
better next question.

\textbf{Scope of the two-rater reading.} Two raters cannot separate agreement between people in
general from agreement between these two, no three-way figure exists, and the pairs come from one suite.
The two are not interchangeable: the second tracks the model panel more closely than the first, $86.7\%$
against $68.5\%$ on the $51$ items both people and the panel decided, $13$ against $4$ on the items
where the two differ. Where the panel was undecided the people split too.
Neither person judged the prerequisite edges.

\subsection{What readers make of the need graphs}

\textbf{Two models read the nodes, on one suite.} On $420$ nodes of 2WikiMultiHopQA shown blind, with
twenty planted foils all caught, two raters of different families agreed on $96.9\%$, a
chance-corrected $0.776$ over three categories, and endorsed $265$ of the $267$ nodes the graph calls
required. A graph calls a need optional when its builder links it to no evidence paragraph, which
happens only on StrategyQA and 2WikiMultiHopQA, and coverage is a ratio over the required needs alone,
so an optional need never enters it. On
2WikiMultiHopQA a need is optional when the builder's paragraph locator finds no paragraph for it, and
of the nodes the graph calls optional there, $96.2\%$ $[91.5, 98.4]$ are really required, so
\textbf{coverage is overstated, in one direction}, on the $3.3\%$ of 2WikiMultiHopQA's nodes
that are optional. On StrategyQA the $18.6\%$ of nodes that are optional are the steps no
annotator linked to a paragraph, operations over earlier answers, so that the same overstatement holds
there is an assumption, not a reading. The reading covers $1.3\%$ of one suite, and the other two
suites have no node-level reading.

\textbf{A rater cannot read an edge, so the edges were corrupted on purpose.} As the benchmark writes
it, the dependent need contains the prerequisite as a literal substring on $200$ of $200$ items, so
every edge reads as necessary. With the prerequisite resolved, a rater rejected six of six true
prerequisites. That instrument returns no value rather than a perfect one. A recomputation of depth over
the stored runs of the run set aside (Appendix~\ref{app:seedzero}) dropped a fifth and then two fifths of the prerequisite edges at random, five draws
each. Depth-weighted recall moves by at most $0.006$ on any suite and never changes sign, and the
deepest need resolved moves by up to $0.044$, toward zero on all three suites, so a random edge error
can hide a depth result and cannot manufacture one. The test covers random deletion only, not a
systematic error or edges that ought to exist and do not, which matter most on StrategyQA, and it is
read at the shared cap.

\subsection{Do the edges gate retrieval? An automated probe}\label{app:edgeprobe}

The criterion was committed before the probe produced any output. For each prerequisite edge in the
held-out tasks, Claude Opus 5, seeing the task and the answers of the parent's ancestors, never the
parent's answer or the child, writes five search queries under each of two prompts, one asking what to
look up next and one asking to answer the task directly. A query hits when the task's own retriever
returns all of the child's evidence in its top five from the released pool. The queries are written
again with the parent's answer added, and with a random same-depth need's answer from another task
instead. An edge is \emph{validated} when the bare question misses the child, at most one query of five
hits without the parent's answer, adding it raises the rate by at least $0.4$ over the better of the two
other contexts, and a model answering the parent from memory fails. It is \emph{refuted} when three or
more queries hit without the parent's answer, or the parent is answerable from memory, and
\emph{untestable} otherwise, including every child with no evidence of its own or evidence shared with
its parent (Table~\ref{tab:app-edgeprobe}).

\begin{table}[h]
\caption{Whether a prerequisite edge gates retrieval: per-query rates at which a query retrieves the child's evidence in the top five, over testable edges, the better of two prompts. StrategyQA parents carry no answer, so their evidence title stands in for it.}
\label{tab:app-edgeprobe}
\begin{center}\small
\begin{tabular}{@{}lccc@{}}
\toprule
 & MuSiQue & StrategyQA & 2WikiMultiHopQA \\
\midrule
prerequisite edges & $466$ & $383$ & $191$ \\
bare task question & $0.35$ & $0.72$ & $0.48$ \\
another task's question (chance) & $0.28$ & $0.24$ & $0.60$ \\
queries without the parent's answer & $0.70$ & $0.72$ & $0.71$ \\
queries with the parent's answer & $0.83$ & $0.72$ & $0.86$ \\
queries with a random entity instead & $0.70$ & $0.74$ & $0.69$ \\
validated / refuted / untestable & $12$ / $391$ / $63$ & $0$ / $119$ / $264$ & $8$ / $144$ / $39$ \\
\bottomrule
\end{tabular}
\end{center}
\end{table}

\noindent The pools are small, twenty paragraphs a task on two suites and ten on the third, and the
retriever returns five, so a child's paragraph is often in reach of any sensible query. The parent's
answer helps on MuSiQue and 2WikiMultiHopQA, where a random entity adds nothing, so the edges carry some
of the dependence they encode. On StrategyQA the parent's evidence title is often already in the
question, and no edge can pass. In these pools an edge records the benchmark's decomposition far more
than a barrier to retrieval, and only $20$ of the $1{,}040$ edges pass the declared test, so the depth
measures credit reaching the evidence a decomposition places deeper, not evidence that could not have
been retrieved before its prerequisite.

With the refuted edges removed and depth recomputed, the trained questioner's depth-weighted recall gain
is $+0.109$, $+0.052$ and $+0.053$ and its deepest-need gain $+0.054$, $+0.043$ and $+0.040$, every one
decided. Graphs that keep only validated edges are nearly flat and cannot separate a depth gain from a
flat one. Eleven of the twelve cells of the two prunings sit inside the band from $200$ random removals
of as many edges per suite and depth, and the twelfth, the StrategyQA deepest need under the first
pruning, falls below its band and sits inside it under the other reading of which class an edge takes
when it both fails the evidence condition and is refuted. On the children themselves, the trained
questioner resolves those of validated edges more often than the same model, prompted, by $+0.205$ on
MuSiQue and $+0.344$ on 2WikiMultiHopQA, over $12$ and $8$ children, and those of refuted edges by
$+0.141$ and $+0.059$, the last with an interval that spans zero, so its gain is not confined to the edges that fail the test.

\subsection{Gold defects, exclusions, and what remains unmeasured}

\textbf{Gold defects and excluded populations.} Twenty tool-use episodes carry a reference reward that
disagrees with the task's own recorded outcome, $4$ airline and $5$ retail on the held-out split and $5$
and $6$ on the training split, all produced by the benchmark's own reference agent
\citep{barres2025tau2}, so no policy's success on those tasks is interpretable and every success figure
excludes them. No corrected release addresses them, and the one for its predecessor
\citep{yao2024taubench,cuadron2025saber} states that the two-agent copies of the same domains had not
been revised. Nineteen episodes minted a second time are excluded by identifier. The $8{,}694$ episodes
produced from an uncommitted working tree, $227$ of them stamped held-out, are admissible as training
data only, because their code cannot be named, which leaves $62{,}185$ scored episodes. The reporting
code's eligibility predicate also requires the held-out split.

\textbf{What remains unmeasured.} Whether a policy can be proactive without harassing its user needs a
precision over questions put to the simulated user, which no scorer computes, so that claim is
unmeasurable with the present code although the episodes exist.
\FloatBarrier
\section{Scale and family}\label{app:scale}

This section gives the scale reading Section~\ref{sec:limits} points to: parameter count does not order
the trained arms, preference training at 32 billion parameters shows no larger ranking gain than at
8 billion, and a second model family was trained by imitation only.
Table~\ref{tab:app-train-ladder} holds the imitation recipe of record fixed across three base sizes
under one scoring version, since comparing each size's strongest checkpoint would attribute a selection
effect to scale.

\begin{table}[ht]
\centering
\caption{The matched-recipe ladder: each cell a paired coverage difference against the same model,
prompted, at gate-matched spend, on the development tasks. Three-seed rows give the mean across
training seeds with the observed range in parentheses, and the largest size is one training run.}
\label{tab:app-train-ladder}
\begin{tabular}{lrll}
\toprule
base & training seeds & MuSiQue ($n=132$) & StrategyQA ($n=333$) \\
\midrule
$1.7$B & $3$ & $+0.1098$ {\scriptsize $(+0.1016, +0.1187)$} & $+0.0309$ {\scriptsize $(+0.0273, +0.0349)$} \\
$4$B   & $3$ & $+0.0434$ {\scriptsize $(+0.0215, +0.0549)$} & $+0.0640$ {\scriptsize $(+0.0611, +0.0691)$} \\
$8$B   & $1$ & $+0.0492$ & $+0.0804$ \\
\bottomrule
\end{tabular}
\end{table}

\textbf{Parameter count does not order these arms.} Every middle-size seed departs from the straight
line between the smallest and largest size, below it on MuSiQue and above it on StrategyQA, each by more
than the seed half-range at that size. The three middle-size seeds are not separable, the comparison's
minimum detectable difference at conventional power being $0.054$, so the replicates settle the
direction of the departure, not its size. A genuine effect of scale does not reverse sign between two
suites on the same checkpoints, so we report no ordering by size, and the ladder is an exploratory
development reading.

\textbf{The prompted comparators stop differently by size.} The smallest model, prompted, never stops at
a covered state, against $0.0066$ and $0.0081$ for the middle size and $0.1525$ and $0.1460$ for the
largest, so a contrast at gate-matched spend pays a checkpoint for stopping well against a comparator that
cannot. Read at each arm's own stop under the shared cap of eight, eleven of the fourteen
checkpoint-by-suite cells behind the ladder go negative: the ladder supports efficiency per question at
gate-matched spend, not more evidence in the end.

\textbf{Where size does show.} On FRAMES, where the trained questioner trails GPT-OSS-120B, prompted,
at the two largest ceilings (Appendix~\ref{app:stopping}), the same model, prompted, trails GPT-OSS-120B
further at every ceiling, with an interval excluding zero from a ceiling of twelve upward, so that
shortfall is size and not the absence of training.

\textbf{Preference training at 32 billion parameters.} The preference stage that fits both pair kinds to
the imitation reference was also run from a 32-billion-parameter imitation checkpoint and read offline on
the development question pairs, as the gain in ranking accuracy over that starting checkpoint, a
development ranking measure and not held-out coverage. Imitation loss rose at every saved checkpoint
after the first, so under a rule fixed before any preference value was read the final checkpoint is
reported first and the lowest-loss checkpoint second (Table~\ref{tab:app-scale-32b-pref}). The final
checkpoint ranks pairs better than its start by $+0.031$ $[+0.001, +0.063]$, better on $24$ of $58$ tasks
and worse on $12$, and the lowest-loss checkpoint by $+0.016$ $[0.000, +0.034]$. Among pairs whose two
questions are within ten tokens of each other in length the final checkpoint's gain is $+0.009$
$[-0.029, +0.044]$, while the 8B arm trained the same way gains $+0.059$ $[+0.017, +0.096]$. At the larger
scale the gain does not survive matching question length and the 8B gain does. The two scales are not
tested against each other.

\begin{table}[ht]
\centering
\caption{Preference training at two scales, read offline on the development question pairs as the gain
in ranking accuracy over each run's own starting checkpoint, with 95\% intervals over tasks. The
32-billion-parameter final checkpoint is reported first by a rule fixed before the reading, because
imitation loss rose at every checkpoint after the first. Length matched keeps the pairs whose two
questions are within ten tokens of each other in length.}
\label{tab:app-scale-32b-pref}
\small
\begin{tabular}{@{}lcc@{}}
\toprule
run & all pairs & length matched \\
\midrule
32B, final checkpoint & $+0.031$ {\scriptsize $[+0.001, +0.063]$} & $+0.009$ {\scriptsize $[-0.029, +0.044]$} \\
32B, lowest imitation loss & $+0.016$ {\scriptsize $[0.000, +0.034]$} & $+0.003$ {\scriptsize $[-0.023, +0.029]$} \\
8B, same recipe & $+0.043$ {\scriptsize $[+0.012, +0.074]$} & $+0.059$ {\scriptsize $[+0.017, +0.096]$} \\
\bottomrule
\end{tabular}
\end{table}

\textbf{A second family, imitation only.} Three Granite 3.3 8B seeds trained with the identical
imitation recipe gain on the held-out split at equal spend against the same model, prompted, by
$+0.038$ to $+0.049$ on MuSiQue and $+0.053$ to $+0.061$ on 2WikiMultiHopQA, every interval excluding
zero. On StrategyQA one seed's interval spans zero and the other two exclude it narrowly, none turning
negative. At each arm's own stop under the shared cap of eight, all three Granite seeds read below the
same model, prompted, on all three suites. Granite has no preference-trained checkpoint, so this is a first-rung reading,
not a family contrast, and a mixture-of-experts base cannot hold the recipe, since its fused experts
leave the adapter four of the seven modules. With every preference pair from one prompted questioner, a
family ladder could show that the recipe carries over, not that it carries over with family-native preference data,
which has not been run.

\FloatBarrier
\section{Prompt search over the questioner}\label{app:gepa}

This appendix reports the prompt search over the questioner to which Section~\ref{sec:limits} points.

\textbf{Two searches found nothing better than the shipped template.} Reflective prompt evolution
\citep{agrawal2025gepa} was run over the questioner's template before anything was trained, scoring
every candidate by rolling out the full episode on multi-hop training tasks. The first search
reported an apparent gain of 87\% on its tie-break slice, but that slice also selected the
winner, and on 24 fresh tasks the gain fell to $+0.027$ at $p = 0.49$. What it had optimized was
length. Across all 25 scored candidates the template ran from 425 to 683 tokens, and only the shipped
template fit the budget within which arms are compared. A length control that keeps the shipped
content and pads a reserved block with inert text to the winner's length reproduced the winner's
entire advantage, the winner minus that control reading $+0.0032$ at $p = 0.96$. The second search
used disjoint search, selection and verdict slices and an adoption rule committed before its winner
existed. On the verdict slice its winner reads $-0.0613$ $[-0.176, +0.054]$ against the shipped
template and $-0.1394$ $[-0.224, -0.055]$ against the length control, winning 3 of 36, in a design
whose minimum detectable difference was $0.1585$. The best prompt here is the one the work started
with, and the headroom is in what the questioner learns to ask rather than in how it is asked to ask.
Two lessons carry beyond this work: a slice that also selects the winner is not held out, and a search
over an arm that will later be compared to ablations must carry the length budget as a constraint
rather than as a check afterward.

\FloatBarrier
\section{The run set aside, and four complete runs}\label{app:seedzero}

The recipe's final preference stage was trained three times. One run, the first registered, resumed
that stage from a checkpoint, and a trainer defect reloaded the stage's starting weights while keeping
its step count, so it trained for about the last tenth of the stage. In the space the adapter adds to
the model its weights sit $0.007$ from the stage's starting point, against $0.080$ for its own last
checkpoint before the resume and for the other two runs. It was set aside on that measurement before
any outcome of the other two was read, and the paper reports the other two. It is also the run the configuration was
selected on: on development it kept stopping when done above the selection floor, which the two
reported runs miss (Appendix~\ref{app:training}).

It behaves like the question-pair arm it started from, and the difference is what the final stage
buys. It asks $2.6$, $1.6$ and $2.0$ questions on the three held-out suites where the trained questioner asks $4.2$,
$3.0$ and $2.3$, finishes in under three questions on $57\%$, $92\%$ and $80\%$ of runs, and asks at an
unfinished state less often than the same model, prompted, by $0.087$, $0.196$ and $0.032$, where the
trained questioner is within $0.07$ of it. Its coverage gain at equal spend is nearly the trained questioner's, $+0.108$, $+0.060$
and $+0.049$, and at the shared cap of eight it loses on StrategyQA, by $-0.099$
$[-0.132, -0.070]$, where the trained questioner does not. On the constructed corpora it lost breadth at the widest
shape, by $-0.213$, which the trained questioner does not.

\textbf{Four complete runs.} The final stage was trained twice more from the start after this run was
set aside, once with the same seed and once with seed 3, from the same starting adapter, preference file
and configuration, and checks declared before either was read held for both: neither resumed, both ran
all $1{,}968$ steps, and their weights sit from the stage's starting point inside the band of $0.07$ to
$0.09$ declared around the two reported runs. Read under the rule and comparator of Table~\ref{tab:heldout}, every
run's coverage interval lies above zero on every suite (Table~\ref{tab:app-fourseeds}). The runs differ
by less than half of each run's own sampling error, no two of them separate, and pooled over the four the
gain is $+0.113$, $+0.070$ and $+0.053$, each interval excluding zero at $50{,}000$ resamples under three
seeds. On the other metrics of Table~\ref{tab:heldout} every run keeps the sign of each cell that table
decides, and the out-of-order rate stays undecided for every run on the two suites where it is undecided
there. The reading is exploratory, declared before either new run finished, and the reported questioner
remains the average of seeds 1 and 2.

\begin{table}[ht]
\centering
\caption{Coverage at equal spend for the four complete runs of the final stage, each against the same
model, prompted, with both arms at the lower of their two question counts, $200$ held-out tasks per
suite. Seeds 1 and 2 are the runs Table~\ref{tab:heldout} averages, and seed 3 and the retrained seed 0
were trained from the start afterwards. Beneath each point is its 95\% interval at $10{,}000$ resamples.
The pooled column averages the four runs within each task. The runs differ by a between-run standard
deviation of $0.005$, $0.007$ and $0.004$ on the three suites, against a mean within-run bootstrap standard
error of $0.015$, $0.017$ and $0.018$.}
\label{tab:app-fourseeds}
\begin{center}\small\setlength{\tabcolsep}{3pt}%
\begin{tabular}{@{}lccccc@{}}
\toprule
 & seed 1 & seed 2 & seed 3 & seed 0, retrained & four runs pooled \\
\midrule
MuSiQue & $+0.108$ & $+0.116$ & $+0.118$ & $+0.109$ & $+0.113$ \\
 & {\scriptsize $[+0.081, +0.141]$} & {\scriptsize $[+0.090, +0.148]$} & {\scriptsize $[+0.092, +0.148]$} & {\scriptsize $[+0.082, +0.139]$} & {\scriptsize $[+0.088, +0.142]$} \\[1pt]
StrategyQA & $+0.064$ & $+0.076$ & $+0.077$ & $+0.064$ & $+0.070$ \\
 & {\scriptsize $[+0.032, +0.099]$} & {\scriptsize $[+0.045, +0.110]$} & {\scriptsize $[+0.046, +0.111]$} & {\scriptsize $[+0.031, +0.099]$} & {\scriptsize $[+0.041, +0.102]$} \\[1pt]
2WikiMultiHopQA & $+0.048$ & $+0.051$ & $+0.057$ & $+0.055$ & $+0.053$ \\
 & {\scriptsize $[+0.015, +0.084]$} & {\scriptsize $[+0.019, +0.085]$} & {\scriptsize $[+0.026, +0.091]$} & {\scriptsize $[+0.021, +0.093]$} & {\scriptsize $[+0.021, +0.087]$} \\[1pt]
\bottomrule
\end{tabular}

\end{center}
\end{table}

\FloatBarrier
\section{Further related work}\label{app:morerelated}

This section extends Section~\ref{sec:related} with work that bears on this paper's design but is not needed
to follow the main text.

\textbf{Asking for missing information.} Preference optimization toward questions of high expected
information gain teaches models to ask informative questions \citep{mazzaccara2024eigdpo}, language models
can be taught to gather information proactively \citep{huang2025gather}, and they can be augmented to ask
clarification questions for retrieval \citep{chi2024clarinet}. Reinforcement learning for active reasoning can
suffer from what \citet{zou2026selflocking} call information self-locking, and agents that explore can still
ignore what they find \citep{englander2026agentsignore}. Long-horizon proactive search has its own benchmark \citep{vibesearch2026}, and ambient agents act on
event streams without waiting for a prompt \citep{langchain2025ambient}. Q\&D differs in its label: each question is credited with the evidence its
continuation actually retrieved.

\textbf{Deep-research and search agents.} Reasoning models have been extended into deep-research agents that
explore the web \citep{li2025webthinker}, agents have been trained by self-play without supervision or external
data \citep{lu2026searchselfplay,zhang2026piplay}, and deep research and everyday online tasks have open
evaluation sandboxes and benchmarks \citep{coelho2025drgym,clawbench2026}. Reinforcement learning with
verifiable rewards may sharpen rather than extend what the base model can reach \citep{yue2025passk}, and
reported gains can reflect contamination \citep{wu2026contamination}, one reason the held-out runs here are
checked against the training split (Appendix~\ref{app:heldout}).

\textbf{Preference optimization with noisy labels.} Labels read from sampled continuations are noisy.
Alternatives to direct preference optimization change its objective
\citep{azar2023ipo,ethayarajh2024kto,meng2024simpo}, and a provably robust variant targets noisy preferences \citep{chowdhury2024rdpo}. We kept standard direct preference optimization
and did not compare these variants.

\textbf{Agreement and evaluation statistics.} Disagreement among annotators can be inherent rather than noise
\citep{pavlick2019inherent,plank2022problem}, which is how we read the raters' low agreement on the better
proactive move (Appendix~\ref{app:validity}), and chance-corrected agreement for such data is surveyed by
\citet{artstein2008inter}. Language-model judges have reliability and bias problems of their own
\citep{yagubyan2026coinflip}. Evaluations of language models need error bars and paired comparisons
\citep{miller2024errorbars}, small improvements call for a paired bootstrap \citep{du2025pairedbootstrap}, and
our decided rule guards against resampling error rather than controlling a false discovery rate
\citep{benjamini1995fdr}. Agentic benchmarks need explicit checks that a task measures what it claims and that
its outcome is scored correctly \citep{zhu2025abc}.

\FloatBarrier
\section{Prompts and the gold firewall}\label{app:prompts}

This appendix supports the rule of Section~\ref{sec:axes} that nothing on the rollout path may read a
need graph, and reproduces the template that is the policy of Section~\ref{sec:questioner}. Four
layers enforce the rule, each catching a different class of leak.

\textbf{An import contract.} A forbidden-import contract, checked on every continuous integration
run, prevents any package on the rollout or training path from importing the gold-reading package.
It catches a reference to gold written directly into code that runs during a rollout.

\textbf{A type wall.} The task view shares no field name and no base class with a gold node, so a
leak by assignment does not typecheck, and its one constructor raises on an unknown field name and on
a known field whose value is not a string, so gold cannot reach a prompt inside a permitted field by
stringification. It catches code that never imports gold but is handed it.

\textbf{A process split.} A rollout worker runs without the environment variable that names the gold
root, so the accessor raises when anything calls it, and that raise is the layer working. It catches
a read attempted at run time by code that got past the first two layers.

\textbf{A marker in one gold string.} The first three layers govern code, and a gold string already
copied into a prompt is not code. So a marker token is planted in one gold string, and every
serialized request is scanned for it inside the client immediately before dispatch, where the
request still exists and has not been sent, since a cached record keeps only a digest of it. A hit voids the run, and a worker
refuses to begin when gold is built but the marker registry cannot be found, because an unarmed
scanner reports nothing, exactly as a clean one does.

\textbf{Coverage of the fourth layer is per suite by design.} The marker rides in the gold answer,
and every question-answering suite carries it, while the five tool-using and transfer suites have no
gold answer and their need text legitimately reaches judges and two ceiling arms, so on those suites
a leak through a string is caught by no layer. What can be bounded there is verbatim echo of secret
need text in recorded replies: on banking and one other of those five suites a scan of $27{,}465{,}239$
characters over $2{,}053$ runs found no match, with the same scanner shown to fire once the secrecy
exclusion is removed, while on airline and retail every need string already occurs in the domain text
the agent is handed, so no content bound is available there, and nothing bounds paraphrase, or gold
entering a prompt, on any suite.

\textbf{The questioner runs one template, rendered fresh at every turn.} There is no separate system
message: the policy renders the whole template from the current state and dispatches it as a single
user turn. The template takes no budget and no cap, and the renderer refuses any placeholder that
would name one, which keeps the policy budget-blind and a prefix of a long rollout exchangeable with
a true short run. The reserved block is the one the length control in Appendix~\ref{app:gepa}
extends.

\begingroup\fontsize{6.8pt}{8.5pt}\selectfont
\begin{verbatim}
You are the Inquirer. A Drafter is preparing an answer to the task below and you interrogate
it, one internal question at a time, until the answer would not change if you asked again.
You are an internal component. You never address the end user and you never write the answer
yourself. Your only output is the next question, or a decision to stop.

TASK
{{question}}

SUITE INSTRUCTIONS
{{instructions}}

EVIDENCE RETRIEVED SO FAR
{{evidence}}

CURRENT DRAFT
{{draft}}

QUESTIONS ALREADY ASKED, AND WHAT CAME BACK
{{history}}

HOW TO CHOOSE
Ask about a need the task requires and the evidence above does not yet resolve. A need is
latent when the task does not name it: it becomes visible only once earlier evidence has
been read. Prefer such a need over one already named in the task statement, and prefer
either over a rephrasing of a question already asked. Name the entity you are asking about
explicitly, because the retriever matches text and cannot resolve a pronoun. Cite in
parent_uids the units whose content made you aware of this need, or leave it empty if the
need is stated in the task itself.

WHEN TO STOP
Stop when every need the task requires is resolved by the evidence above, or when the next
question you can think of would return something the evidence already contains. Stopping is
a decision on the same footing as asking, not a failure.

{{user_channel}}

RESERVED
This block is reserved. It appears in every condition so that the amount of text surrounding the
task is the same in each. It states no requirement and expresses no preference. It carries no
fact about the task, about the evidence or about the answer. It is not a hint, and nothing in it
should be treated as one. It describes neither what to ask nor when to stop. It names no entity
and refers to no document.

OUTPUT
Reply with exactly one JSON object and no other text, no markdown fence, no commentary:
{"action": "ask", "question": "<a single question>", "rationale": "<one clause>",
 "parent_uids": ["<uid>"]}
or
{"action": "stop", "rationale": "<one clause>"}
\end{verbatim}
\endgroup

\noindent The template opens with ``You are the Inquirer'' (Inquirer is the template's name for the
questioner, and Drafter its name for the drafter). The output contract is one line in the file, wrapped here at the comma before
\texttt{parent\_uids} to fit the column, and nothing else in the block is altered. The user-channel
hole takes a placebo fragment in every arm but one, so opening the channel changes what the prompt
says and not how long it is.

\begingroup\fontsize{6.8pt}{8.5pt}\selectfont
\begin{verbatim}
A NOTE ON THIS SECTION
This paragraph is reserved. It appears in every one of the conditions so that the surrounding
text has the same size in each, and it carries no instruction, no preference and no
information about the task, the evidence or the answer.
\end{verbatim}
\endgroup

\begingroup\fontsize{6.8pt}{8.5pt}\selectfont
\begin{verbatim}
THE USER CHANNEL IS OPEN
You may address one question to the user instead of the knowledge base by setting
"target": "user". Use it only for a need no document can resolve, such as a preference or a
private constraint held by the user alone.
\end{verbatim}
\endgroup
\section{Terms, common questions and released resources}
\label{app:glance}

This appendix gathers the paper's terms in one place, answers the questions a reader is likely to
bring to it, each with a pointer to the full account, and lists what is released.

\paragraph{Keywords.} Proactive agents, LLM agents, information seeking, question asking, clarifying
questions, need graphs, multi-hop question answering, agentic search, retrieval-augmented generation,
preference optimization, customer-service agents, $\tau^2$-bench.

\subsection{Terms}

\begin{description}
\item[Content of proactivity.] What an agent pursues without being asked, as distinct from whether
and when it acts on its own (Section~\ref{sec:intro}).
\item[Frontier.] The needs an agent can already name at a point of a run but has not yet resolved.
\item[Horizontal proactivity.] Pursuing a need on the frontier other than one the need just resolved
made nameable: a need the request implied, like a customer's account looked up from the name and ZIP
code the customer gave, or one that earlier evidence opened, like another order in the same account
(Section~\ref{sec:axes}).
\item[Vertical proactivity.] Pursuing a need that the need just resolved made nameable, as the
account's record names the order holding the boots (Section~\ref{sec:axes}).
\item[Need graph.] For each task, the needs, the units of evidence the task requires, with a
prerequisite edge wherever one need can be named only after another is resolved. It is recovered
mechanically from the decompositions the benchmarks ship. It scores finished runs and labels training
data, and it is never shown to the agent.
\item[Depth.] The length of the longest prerequisite path above a need. Needs at depth zero can be
named from the request, and deeper ones only after the evidence above them.
\item[Required-evidence coverage.] The share of a task's required needs that a run recovers. It reads
both forms of proactivity.
\item[Breadth and depth-weighted recall.] Breadth counts the independent lines of inquiry a run
advances past their first need, and reads horizontal proactivity. Depth-weighted recall averages the
coverage at each depth with weights that favor deeper levels, and reads vertical proactivity
(Table~\ref{tab:defs}).
\item[Equal retrieval spend.] Policies are compared after the same number of questions, so asking
more cannot pass for asking better.
\item[Questioner, drafter and answerer.] The questioner asks one question at a time or stops, a
retriever answers it from the task's evidence pool, a frozen drafter keeps the draft, and a frozen
answerer writes the final answer. Only the questioner differs between the policies compared.
\item[Q\&D (questioner and drafter).] The training algorithm this paper proposes. It forks a recorded
run, continues it after several candidate questions, and prefers the question whose continuation
retrieves more of the required evidence, with no reward model or judge (Section~\ref{sec:questioner}).
\end{description}

\subsection{Common questions}

\paragraph{What does it mean for an LLM agent to be proactive about content?}
A tool-using LLM agent usually does what the user asks, yet completing the task often needs
information the user never mentions. Work on proactive agents mostly studies whether and when an agent
should act on its own \citep{horvitz1999mixed,lu2024proactivebench,tang2026proagentbench}. This paper studies what the agent
pursues unasked, in a horizontal and a vertical form, and measures and trains both.

\paragraph{How is proactivity measured without an LLM judge?}
Each task carries a need graph recovered from the benchmark's own decomposition of its question. A
finished transcript is scored by which required evidence it retrieved, matched by identifier, at which
depth, and whether the agent stopped at the right time (Section~\ref{sec:axes},
Appendix~\ref{app:metrics}). No quantity the paper reports is judged by a model.

\paragraph{How does Q\&D train a questioner without a reward model?}
From the consequences of its questions. A recorded run is forked at a step, eight alternative
questions are sampled there, and each is continued to the end. Pairs are ordered by whether the run
answered the task, then by which reached the complete evidence sooner, then by how much evidence each
turn added. The questioner is trained in three stages: imitation of good decisions, direct preference
optimization \citep{rafailov2023dpo} on question pairs, and direct preference optimization on question
pairs together with contrasts that rank asking above stopping at unfinished states
(Section~\ref{sec:questioner}, Appendix~\ref{app:training}).

\paragraph{How does this differ from training agentic search on answer correctness?}
Search agents trained with outcome rewards \citep{jin2025searchr1} credit a whole run with whether its
final answer was right. Q\&D credits each question with what followed it, so a question that reaches an
unstated need early wins even when the final answers agree. Answers decide only 6\% of the pairs Q\&D
trains on.

\paragraph{How does this differ from asking clarifying questions?}
Work on clarifying questions learns what to ask the user when a request is ambiguous or incomplete
\citep{rao2018learning,aliannejadi2019asking}. Here the questioner goes after the missing information
itself, in the evidence it can retrieve. In a customer-service agent it completes more retail tasks
than a prompted model $15\times$ larger with fewer follow-up turns from the customer, which the paper
attributes to the agent finding more of what the task needs in the store's records
(Appendix~\ref{app:transfer}).

\paragraph{Which benchmarks and models does the paper use?}
Training data are mined from MuSiQue \citep{trivedi2022musique}, StrategyQA \citep{geva2021aristotle}
and 2WikiMultiHopQA \citep{ho2020twowiki}, and results are read on held-out test splits of 200 tasks per
benchmark. The trained questioner is Qwen3-8B \citep{yang2025qwen3}. GPT-OSS-120B
\citep{openai2025gptoss} is the drafter and the answerer in every arm except the single-model control,
and a prompted comparator in the questioner's role. The transfer study uses the retail and airline
domains of $\tau^2$-bench \citep{barres2025tau2} with a simulated customer (Section~\ref{sec:setup}).

\paragraph{What are the main results?}
At equal retrieval spend, the trained 8B questioner recovers more of the required evidence than the
same model, prompted, on every suite, and reaches deeper into each task's dependencies. On MuSiQue it
recovers 90\% of the required evidence against 78\%. It leads GPT-OSS-120B, a prompted model
$15\times$ larger in the same role, on MuSiQue and StrategyQA, by 7.0 and 4.3 points, and trails it on
2WikiMultiHopQA, by 3.5 points. The gain comes from what it asks, not from asking more or longer
questions. Stopping on its own, it asks fewer questions than either prompted model and still leads on
the deepest chains (Section~\ref{sec:results}).

\paragraph{Does the questioner work beyond question answering?}
Yes, without further training. Placed in a customer-service agent on $\tau^2$-bench, it raises retail
task success from 13\% and 12\% to 34\% and 32\% under two prompt variants, better on ten of
twenty-five tasks and worse on none in each. It asks less and finds more: 1.3 and 1.4 fewer questions
per retail dialogue, more of which reach the records the task needs. In airline, task success rises by
6.9 and 6.4 points. Against GPT-OSS-120B in retail, it completes 17.0
and 12.7 points more tasks, with 1.6 and 1.8 fewer follow-up turns from the customer
(Appendix~\ref{app:transfer}).

\paragraph{Does finding more evidence give better final answers?}
Not yet detectably. The extra evidence does not yet reach the final answers, and training teaches what
to ask more readily than when to stop (Section~\ref{sec:limits}, Appendix~\ref{app:answerloss}).

\paragraph{What are the limitations?}
The two reported seeds share one supervised checkpoint and one pair export, so intervals are over
tasks alone. The released retrieval pools are small enough that most prerequisite edges do not gate
retrieval, so the vertical result concerns depth in a decomposition. Every user-facing number comes
from a simulated customer, and Q\&D is one round of off-policy training, not compared with on-policy
reinforcement learning (Section~\ref{sec:limits}).

\paragraph{Where are the code and the trained model?}
The project page is \url{https://dolev31.github.io/ProactiveInquirer/}. The code, with the agent loop,
the need-graph builders and metrics, the training ladder, the analyses and a test suite, is at
\url{https://github.com/dolev31/ProactiveInquirer}. The trained questioner is at
\url{https://huggingface.co/dolev31/ProactiveInquirer-Qwen3-8B}, a LoRA adapter with both training
seeds, with merged weights and GGUF quantizations in companion repositories.

\subsection{Where this work sits}

The paper connects several lines of work, discussed in Section~\ref{sec:related} and
Appendix~\ref{app:morerelated}: proactive and mixed-initiative agents
\citep{horvitz1999mixed,handler2023taxonomy,lu2024proactivebench}, benchmarks that test whether a model
recognizes missing information or unstated needs \citep{li2025questbench,harfi2026proactbench},
learning what to ask, including clarifying questions
\citep{rao2018learning,aliannejadi2019asking,andukuri2024stargate,chen2025act}, agentic search and
retrieval-augmented generation
\citep{yao2023react,press2023selfask,trivedi2023ircot,asai2024selfrag,jin2025searchr1}, question
decomposition for multi-hop question answering \citep{wolfson2020break,zhou2023leasttomost}, step-level
preference training \citep{wang2024mathshepherd,lai2024stepdpo}, and simulated users for evaluating
agents \citep{qian2025userbench,barres2025tau2}.

\end{document}